\documentclass[10pt,journal,compsoc]{IEEEtran}
\usepackage{hyperref}
\usepackage{graphicx}
\usepackage{subfigure}
\usepackage[flushleft]{threeparttable}
\usepackage{booktabs} 
\usepackage{amsfonts}
\usepackage{multirow}
\usepackage{makecell}
\usepackage{tabularx}
\usepackage{xcolor}
\usepackage{footnote}
\usepackage{caption}
\usepackage{textcomp, gensymb}
\usepackage{amsmath}

\usepackage{algorithm}
\usepackage{algpseudocode}

\newcolumntype{M}{>{$}c<{$}}
\newcolumntype{Z}{>{\centering\arraybackslash}X}
\newcolumntype{C}{>{\centering\arraybackslash}p}
\newcolumntype{Y}{>{\centering\arraybackslash}X}
\newcolumntype{L}{>{\raggedright\arraybackslash}p}

\newcommand{\rowstretch}[1]{\renewcommand{\arraystretch}{#1}}

\ifCLASSOPTIONcompsoc
  \usepackage[nocompress]{cite}
\else
  \usepackage{cite}
\fi

\ifCLASSINFOpdf
\else
\fi

\usepackage{amsmath}
\begin{document}

\title{Universal Fingerprint Generation: Controllable Diffusion Model with Multimodal Conditions}
\author{Steven~A.~Grosz, and~Anil~K.~Jain,~\IEEEmembership{Life~Fellow,~IEEE}
\IEEEcompsocitemizethanks{\IEEEcompsocthanksitem S.A. Grosz and A.K. Jain are with the Department of Computer Science and Engineering, Michigan State University, East Lansing, MI, 48824 USA (e-mail: groszste@cse.msu.edu, jain@cse.msu.edu).}
}

\markboth{}{Grosz and Jain: Universal Fingerprint Generation: Controllable Diffusion Model with Multimodal Conditions}

\IEEEtitleabstractindextext{%
\begin{abstract}
    The utilization of synthetic data for fingerprint recognition has garnered increased attention due to its potential to alleviate privacy concerns surrounding sensitive biometric data. However, current methods for generating fingerprints have limitations in creating impressions of the same finger with useful intra-class variations. To tackle this challenge, we present \textit{GenPrint}, a framework to produce fingerprint images of various types while maintaining identity and offering humanly understandable control over different appearance factors such as fingerprint class, acquisition type, sensor device, and quality level. Unlike previous fingerprint generation approaches, GenPrint is not confined to replicating style characteristics from the training dataset alone: it enables the generation of novel styles from unseen devices without requiring additional fine-tuning. To accomplish these objectives, we developed GenPrint using latent diffusion models with multimodal conditions (text and image) for consistent generation of style and identity. Our experiments leverage a variety of publicly available datasets for training and evaluation. Results demonstrate the benefits of GenPrint in terms of identity preservation, explainable control, and universality of generated images. Importantly, the GenPrint-generated images yield comparable or even superior accuracy to models trained solely on real data and further enhances performance when augmenting the diversity of existing real fingerprint datasets.
\end{abstract}

\begin{IEEEkeywords}
Artificial Fingerprint Generation, Synthetic Fingerprints, Denoising Diffusion Probabilistic Models, Latent Diffusion Models, Zero-shot Image Generation.
\end{IEEEkeywords}}

\maketitle
\IEEEdisplaynontitleabstractindextext

\begin{figure*}
\includegraphics[width=\textwidth]{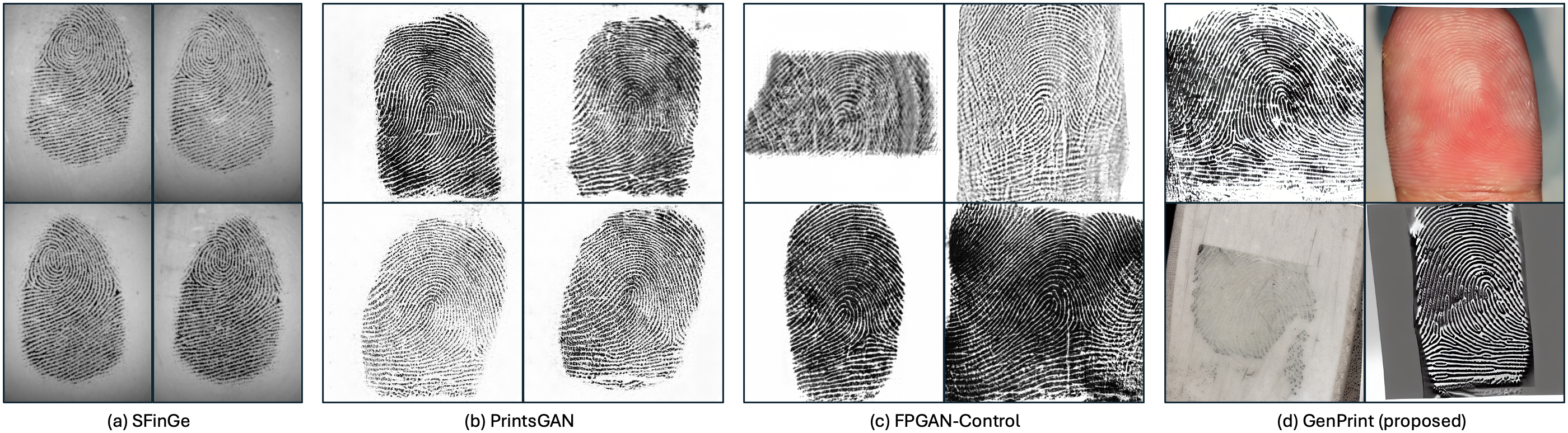} 
\caption{Synthetic fingerprint images generated by various baseline methods and the proposed GenPrint. The four images in each panel are impressions of the same finger to show case the intra-class variance of each method.}
\label{fig:baseline_imgs}
\end{figure*}

\IEEEraisesectionheading{\section{Introduction}\label{sec:introduction}}
\IEEEPARstart{T}{he} use of Artificial Intelligence Generated Content (AIGC) over the last few years has exploded due to advancements in model architectures and larger computation and data being used to train Generative AI (GAI) models~\cite{cao2023comprehensive}. In particular, text generation models, such as ChatGPT, have catapulted the field of GAI into the public view since its public release in November of 2022~\cite{openai2022chatgpt}. Following in its wake came stunning advancements in image and video generation models, such as ImageGen~\cite{google-imagen} and SORA~\cite{sora_openai}, utilizing denoising diffusion probabilistic model (DDPM) frameworks. Since then, DDPM models have proliferated as the center of attention in many top computer vision conferences and journals. Notably, their probabilistic framework and straight-forward optimization process makes DDPMs more stable and easy to train compared to generative adversarial networks (GANs)~\cite{goodfellow2014generative}, one of the predominant frameworks for image generation previously. Furthermore, the work of Dhariwal and Nichol further demonstrated the advantages of diffusion models over GANs for image generation in terms of image quality~\cite{dhariwal2021diffusion}. Indeed, the introduction of GANs by Goodfellow et al.~\cite{goodfellow2014generative} in 2014 and the recent surge in DDPM models have revolutionized GenAI capabilities across enumerable industries and applications.

Artificial fingerprint generation is one application which has received increased interest for the potential of synthetic data for training and evaluation of algorithms, aided by recent privacy and ethical concerns as well as difficulty and cost associated with collecting biometric data. Before the explosion of deep learning techniques, fingerprint generation methods began with intelligent, hand-crafted methods to simulate convincing fingerprint patterns and textures~\cite{cappelli2004sfinge}. Importantly, these methods allowed for generating multiple images of the same finger, opening the door to training and evaluation of fingerprint recognition algorithms.

Early GAN-based methods drastically improved the realism of the generated prints but lacked control over the fingerprint identity being generated~\cite{mistry2019fingerprint,finger-gan,synfi,lightweight,bahmani2021high,cao2018fingerprint}. Subsequent works aimed to fill this gap by replacing each stage of the mutli-stage generation pipeline of hand-crafted methods with GANs, preserving the identity of the generated fingerprints at each stage~\cite{level-3, engelsma2022printsgan, grosz2022spoofgan}. However, with the exception of identity, other appearance factors remained obscured and uncontrollable, such as the specific fingerprint class (e.g., arch, loop, and whorl), acquisition type (e.g., rolled, slap, contactless, swipe, and latent), sensor characteristics (e.g., optical, capacitive, thermal, etc.), and quality level (e.g., high, average, and low) of the generated prints. Shoshan et al.~\cite{shoshan2024fpgan} proposed FPGAN-Control to disentangle identity and appearance factors in the latent space and allowed for swapping between different appearance latent vectors to achieve some degree of control over intra-class variations (e.g., acquisition type, sensor, and pressure level); however, this method lacked explicit, humanly explainable control over appearance factors.

Recent advancements in text to image generation models utilizing DDPMs have demonstrated very realistic and controlled image generation capabilities. In this work, we aim to leverage DDPM advancements for controllable fingerprint image generation utilizing multimodal conditions (text and image) for improved generation capabilities. We leverage text prompts to allow for guidance of explainable appearance factors and rely on image style embeddings for factors not easily expressed in language. Importantly, an added benefit of our novel image style condition is that the generation outputs are no longer constrained to interpolating between the domain of the seen training data, for it allows for zero-shot generation of novel fingerprint sensor characteristics not seen during training. For a visual comparison, figure~\ref{fig:baseline_imgs} shows some example synthetic images generated from SFinGe, PrintsGAN, FPGAN-Control and the proposed model which we refer to as GenPrint. The four images in each panel are of impressions of the same finger identity to showcase the intra-class variance of each method, which demonstrates the improved diversity which GenPrint is capable of generating.

More concisely, the contributions of this research are the following:

\begin{enumerate}
    \item A controllable latent diffusion model, GenPrint, using text and image conditions for highly realistic and diverse synthetic fingerprint generation.
    \item GenPrint is capable of generating fingerprints of any acquisition type, sensor, fingerprint class, and quality, including fingerprint styles not seen during training without any additional fine-tuning (e.g., zero-shot fingerprint style generation).
    \item The generation process is controllable (both in appearance and identity preservation) and explainable with humanly interpretable text prompts.
    \item The utility of GenPrint synthetic images is validated through experiments showcasing improved recognition performance of models trained on GenPrint images compared to real datasets and other benchmark fingerprint generation methods.
    \item We also demonstrate the utility of GenPrint images for evaluating fingerprint recognition systems by replacing real data for large-scale identification experiments.
    \item Upon acceptance and publication, a dataset of 100K synthetic finger identities with 15 impressions of various acquisition devices will be released to the research community.
\end{enumerate}

\begin{figure*}
\includegraphics[width=\linewidth]{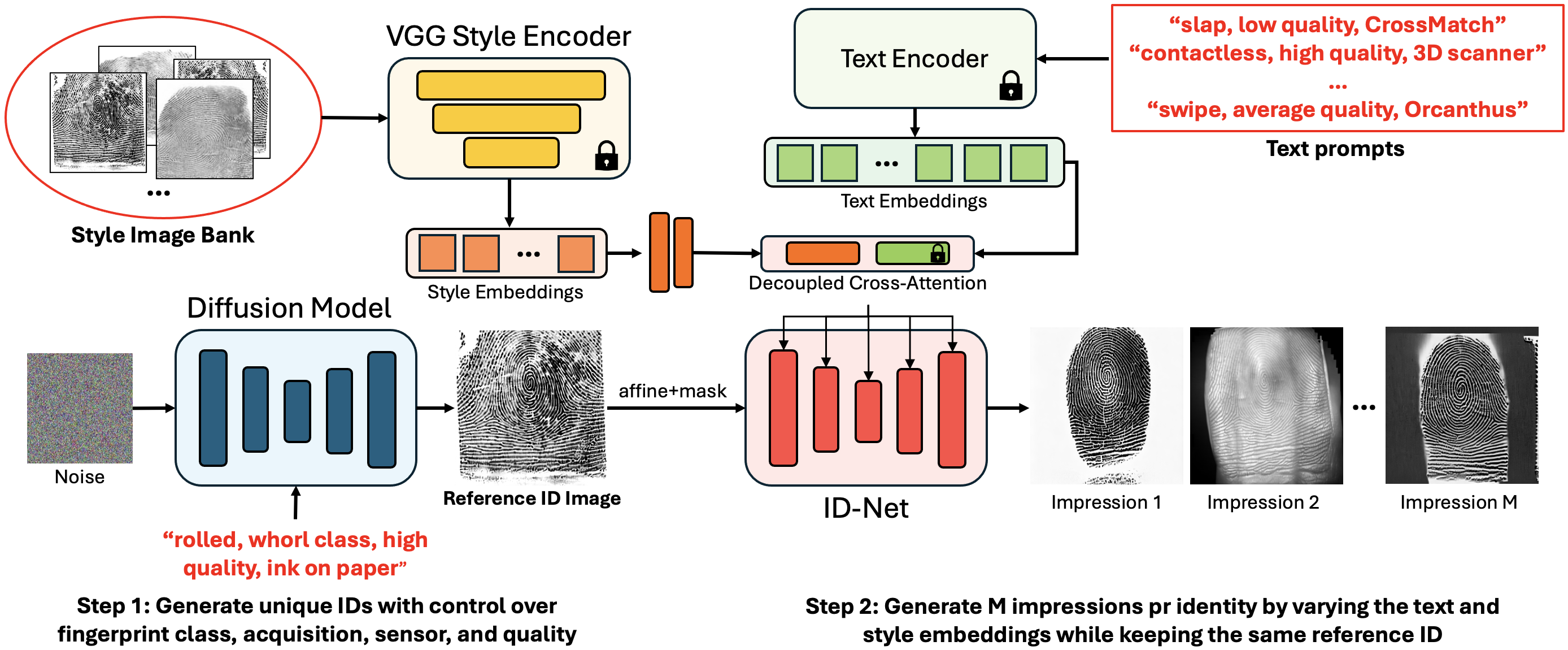} 
\caption{Architecture of GenPrint.}
\label{fig:architecture}
\end{figure*}

\section{Related Work}
\subsection{Hand-crafted Fingerprint Generation Methods}
The seminal work of Cappelli et al.\cite{cappelli2004sfinge} utilized a combination of an elliptical shape generation model, mathematical ridge flow and Gabor filters for ridge pattern generation, and noise and distortion models to simulate realistic fingerprint patterns. Importantly, this model allowed for generating multiple impressions of the same finger leading to its adoption for aiding in training and evaluation of fingerprint recognition models. Despite its impressive capabilities and intelligent design, SFinGe is limited in its intra-class variations it is able to generate due to its hand-crafted nature (see subfigure (a) of Figure~\ref{fig:baseline_imgs} for examples). More recent methods have turned to deep learning techniques, starting with GANs, to learn the subtle intra-class variations that have led to more varied and realistic fingerprint images. 

\subsection{Fingerprint Generation via GANs}
The introduction of GANs gave way to more realistic fingerprint generation that captured more realistic texture characteristics that are difficult to hand-design~\cite{mistry2019fingerprint,finger-gan,synfi,lightweight,bahmani2021high,cao2018fingerprint}; however, early uses of GANs lacked control over the fingerprint identity being generated - severely limiting the utility of the generated fingerprints. Wyzykowski et al.~\cite{level-3}, aimed to fill this gap by adopting CycleGAN as a wrapper around SFinGe generated images to impart them with more realistic textures, while leveraging SFinGe's ability to generate multiple impressions. However, the intra-class and inter-class variations were still limited by the hand-designed generation of SFinGe. Engelsma et al. went one step further and designed a multi-stage GAN method for generating highly realistic fingerprint ridge patterns with multiple impressions per finger and showed substantial improvement over SFinGe in utility for recognition model training~\cite{engelsma2022printsgan}. Finally, Shoshan et al.~\cite{shoshan2024fpgan} adopted a mixed variational autoencoder (VAE) and GAN architecture called FPGAN-Control to interpolate between latent identity and appearance vectors to be able to render fingerprint images in multiple different appearances. Still, this model lacked explicit control over the appearance factors, and the possible space of generated fingerprint styles is constrained to the distribution of styles belonging to the original training set. 

\subsection{DDPMs for Fingerprint Generation}
To the best of our knowledge, DDPMs have only just begun to be investigated for artificial fingerprint generation~\cite{li2023diffusion, tang2024enhancing}. Tang et al. applied a vanilla DDPM to synthesize unconditional fingerprint patches and validated the realism compared to real fingerprint patches using the Fréchet Inception Distance (FID) metric~\cite{tang2024enhancing}. Li and Yang also applied an unconditional DDPM model trained on a dataset of latent, rolled, and plain (i.e., slap) fingerprint images to randomly generate fingerprint impressions of these types~\cite{li2023diffusion}. They demonstrated the realism of the DDPM generated fingerprint images both in terms of NFIQ quantitative values and t-SNE qualitative comparisons to the real fingerprint images. However, their model lacked control over both the identity and appearance of the generated fingerprints, which is critical for training and evaluation of fingerprint recognition models. To the best of our knowledge, the proposed GenPrint model is the first use of DDPMs for fingerprint generation with explicit control over both the identity and appearance of generated fingerprint images.

\section{GenPrint: Controllable Multimodal Fingerprint Diffusion Model}
GenPrint is a multimodal latent diffusion model~\cite{rombach2022high} finetuned for fingerprint generation from a pretrained Stable Diffusion v1.5 model with weights made available from the Diffusers library~\cite{von-platen-etal-2022-diffusers}. In this section, we first describe the text to image fingerprint generation capabilities including the dataset curation process and fine-tuning procedure. Next, we describe the architectural design for incorporating style image embeddings into the Stable Diffusion pipeline and explain the zero-shot style generation capability it facilitates. Finally, the identity preservation process is described along with a detailed description of the full pipeline for generating synthetic fingerprint images with GenPrint. An overview of the architecture design is given in Figure~\ref{fig:architecture}.

\subsection{Control Factors via Text Conditions}
The first step in fine-tuning Stable Diffusion for text to fingerprint generation is obtaining a large corpus of fingerprint images and associated text descriptions. For this purpose, we aggregated data from multiple fingerprint datasets from predominately publicly available sources. These training datasets are listed in Table~\ref{tab:datasets} along with the acquisition label, sensor label, and number of images for each dataset. Our aggregated dataset consists of data from five different acquisition types (rolled, slap, swipe, contactless, and latent) and thirty different sensing devices ranging from optical readers as well as capacitive, thermal, contactless, and latent surfaces.

Missing from many fingerprint datasets are annotations for fingerprint class (whorl, plain arch, tented arch, left loop, and right loop) and quality (low, average, and high quality), which are needed to impart the generator with this kind of control. To obtain these labels, we utilized Verifinger SDK v12.4\footnote{\url{https://www.neurotechnology.com/verifinger.html}} to extract class and NFIQ 2.0~\cite{tabassi2021nfiq} quality estimations. Since the NFIQ 2.0 metric was optimized for slap impressions utilizing frustrated total internal reflection (FTIR) optical imaging, the quality levels across each acquisition type may vary distinctly. Thus, we fit individual quality distributions according to a normal distribution using images belonging to each acquisition category and assigned low, average, and high quality labels to image clusters based on the mean $\pm$ standard deviation.

Using these annotations we constructed text prompt labels for each training image utilizing the following template: ``a \{acquisition\} fingerprint image, \{class\} pattern, \{quality\} quality, \{sensor\}, \{sensing\}", where
the acquisition type is one of \{rolled, slap, swipe, contactless, latent\}, class is one of \{whorl, plain arch, tented arch, left loop, right loop\}, quality is one of \{low, average, high\}, sensor is one of the thirty training sensors listed in Table~\ref{tab:datasets}, and sensing type is one of \{FTIR optical, direct-view optical, multispectral optical, capacitive, thermal\}.

For fine-tuning Stable Diffusion on our text to fingerprint image dataset, we utilize the low-rank adaptation (LoRA) strategy for more efficient training with a rank of 128~\cite{hu2022lora}. The LoRA weights are finetuned with a learning rate of 0.0001, cosine scheduler~\cite{loshchilov2016sgdr}, default Adam optimizer~\cite{kingma2014adam}, and batch size of 96 spread across 8 Nvidia A100 GPUs. The model is trained for 500,000 steps and trained on fingerprint images of a resolution of $512\times512$ pixels.

\begin{table*}
\caption{Training datasets for GenPrint.}
\label{tab:datasets}
\rowstretch{1.5} 
\begin{tabularx}{\textwidth}{>{\centering\arraybackslash}p{0.2\linewidth}|>{\centering\arraybackslash}p{0.1\linewidth}|>{\centering\arraybackslash}p{0.45\linewidth}|>{\centering\arraybackslash}p{0.15\linewidth}}
\toprule[1pt]
\textbf{Train Dataset} & \textbf{Acquisition Types} & \textbf{Sensor Types} & \textbf{No. Images (Fingers)} \\
\Xhline{1pt}
NIST SD14~\cite{sd14} & Rolled & Ink on paper & 54,000 (27,000) \\
\hline
FVC 2002~\cite{fvc2002} & Slap & Desktop Scanner, TouchChip, DF90 & 2,400 (100) \\
\hline
FVC 2004~\cite{fvc2004} & Slap, Swipe & CrossMatch, Digital Persona, Fingerchip & 2,400 (100) \\
\hline
PLUS-MSL-FP~\cite{kirchgasser2021plus} & Slap, Swipe & Eikon, Integrated Biometrics Columbo, Integrated Biometrics Curve, Lumidigm, Next Biometrics, Suprema RealScan G1, Digital Persona & 106,712 (580) \\
\hline
MSU Infant Fingerprint~\cite{j2019infant} & Slap & SilkID & 9,683 (1,921) \\
\hline
NIST SD302 (N2N)~\cite{nist302} & Slap, Contactless, Rolled & CrossMatch, Eikon, GreenBit, ANDI, S120, MorphoWave, DactyScan, LIVETOUCH, Futronic, RaspiReader & 45,072 (1,600) \\
\hline
NIST SD302 Latent~\cite{nist302} & Rolled, Latent & Ink on paper, crime scene & 7,586 (1,019) \\
\hline
MSP Latent~\cite{yoon2015longitudinal} & Rolled, Latent & Ink on paper, crime scene & 1,866 (933) \\
\hline
IIITD SLF~\cite{sankaran2012hierarchical} & Slap, Latent & CrossMatch, crime scene & 480 (150) \\
\hline
MOLF~\cite{sankaran2015multisensor} & Slap, Latent & Lumidigm, Secugen, CrossMatch, crime scene & 65,512 (1,000) \\
\hline
MUST~\cite{malhotra2023multi} & Slap, Latent & CrossMatch, crime scene & 20,247 (120) \\
\hline
IIT Bombay Touchless and Touch-based~\cite{birajadar2019towards} & Slap, Contactless & eNBioScan, smartphone & 3,200 (200) \\
\hline
ISPFDv2~\cite{malhotra2020matching} & Slap, Contactless & Secugen, smartphone & 57,600 (304) \\
\hline
UWA Benchmark 3D Fingerprint~\cite{zhou2014benchmark} & Slap, Contactless & CrossMatch, 3D scanner & 18,266 (1,500) \\
\hline
ZJU Finger Photo and Touch-based Fingerprint~\cite{grosz2021c2cl} & Slap, Contactless & Digital Persona, smartphone & 39,580 (824) \\
\bottomrule[1pt]
\end{tabularx}
\end{table*}

\subsection{Zero-shot Style Generation}
Motivated by the fact that many of the textural intra-class variations present in fingerprint images are not easily expressed in language via simple text prompts, we turned toward a deep learning-based representation to capture those characteristics. In particular, we take a pretrained VGG~\cite{simonyan2014very} model trained on ImageNet to embed style embeddings for each training image. These style embeddings are injected into the diffusion model via cross-attention layers which are de-coupled from the cross-attention layers from the textual embeddings used to control the explainable style factors. Our choice of VGG embeddings for style representation is motivated from two key insights: i.) the previous use of VGG for neural style transfer~\cite{johnson2016perceptual} and ii.) visualizing the separation of VGG style embeddings for various fingerprint sensor types in the t-SNE embedding space (see Figure~\ref{fig:tsne}).

During inference, style embeddings from various sensor types present in the training data can be sampled to generate images of that sensor. On the other-hand, even style embeddings extracted from images of a completely new, unseen sensor can be used to generate images in that new sensor domain. Therefore, our method is generalizable and allows for ``zero-shot" fingerprint style generations without any additional fine-tuning required. This fact is later supported by empirical evidence in section~\ref{sec:zeroshot} to produce new fingerprint characteristics of latent, optical, capacitive, and contactless sensors outside those seen during training. 

\subsection{Fingerprint Identity Preservation}
Several strategies for identity preservation and personalization in diffusion models have been proposed. Some of these techniques, such as Textual Inversion~\cite{gal2022image} and DreamBooth~\cite{ruiz2023dreambooth}, require additional fine-tuning for each new concept, whereas others, such as IP-Adapter~\cite{wang2024instantid} and PhotoMaker~\cite{li2023photomaker}, can produce identity consistent generations for multiple subjects without inference time fine-tuning. Both IP-Adapter and PhotoMaker embed the identity of an input reference image or images into the diffusion process via cross-attention layers. This guides the diffusion model to generate images which are identity consistent with the input reference images. Empirically, we tried IP-Adapter but found that it lacked the fine-grained spatial control needed to maintain the fingerprint ridge structure throughout the image. To solve this, we turned to ControlNet~\cite{zhang2023adding}, which is another adaptation to the diffusion model process in which reference images are provided to the diffusion model to guide the generation with spatially consistent outputs.

For fingerprints, the identity discriminative features which are consistent across multiple different acquisition and sensor types are the silhouettes of the ridge flow patterns giving rise to the relative orientation of minutiae points of each finger. We posit that ControlNet is a suitable choice for imparting our DDPM model with identity preservation. Therefore, we propose to adapt the ControlNet framework to provide explicit spatial consistency of the generated fingerprint ridge pattern by pre-pending a ridge extraction module to the input of our identity preserving diffusion model, ID-Net. This ridge extractor removes sensor dependent and other style characteristics from the input fingerprint control image leaving only the ridge pattern silhouette image to guide the spatial preservation of the fingerprint identity, including the location and orientation of minutiae points. This, combined with the text and style embeddings providing the style information, allows our ID-Net to generate varying textural characteristics while maintaining the input fingerprint ridge pattern. The architecture for our ridge extraction model is the light-weight SqueezeUNet model, which has been successfully applied previously for fingerprint ridge extraction~\cite{grosz2023latent}. 

\subsection{Generation Pipeline for GenPrint}
The full generation pipeline for GenPrint consists of two stages. First, our finetuned stable diffusion model is used to generate full (i.e., rolled) fingerprint images of various fingerprint classes from a random noise vector. For this stage, the text prompt guiding the generation follows the template of ``a rolled fingerprint image, \{class\} pattern", high quality, ink on stock paper", where the fingerprint class is randomly selected from the five available classes. This provides a full fingerprint ridge pattern for use in the subsequent generation stage which imparts controllable style variations to generate large intra-class variations. By varying the noise vector for each generation, completely new and unique fingerprint patterns are generated. This fact is supported in section~\ref{sec:capacity}, showcasing the inter-class separation of the generated fingerprints, and section~\ref{sec:leakage}, highlighting the low similarity between generated identities and the training fingerprint identities. In the second stage, the generated fingerprint images from the first stage are passed through ID-Net and imparted with varying appearances based on the style embeddings (from reference images either belonging to the training set or from new example images from unseen sensors) and different text prompts providing explainable acquisition, sensor, and quality factors.

One critical observation we noticed was that the ControlNet framework was indeed very successful at constraining the local spatial details to be preserved in the generated images; however, often we found that the generated images had the tendency to over-constrain the generation process to preserve every detail of the input control image. This is undesirable if, for example, the desired output image is a slap fingerprint image where the input control image is a full, rolled fingerprint ridge pattern. The result is an unrealistic image with the full rolled fingerprint pattern in the style of the specified ``slap" sensor input. Therefore, we apply a mask to the output of the ridge extractor aligned with the input text prompt to apply a realistic foreground mask for the specified acquisition type. For example, if the prompt is to produce a slap fingerprint image, then an extracted mask of the fingerprint foreground area from one of the slap training images is applied to the input rolled fingerprint ridge pattern to produce an image with a fingerprint area resembling a realistic slap fingerprint. If instead the prompt is to generate a latent fingerprint, then a mask from a training latent fingerprint image is applied to the input control image to produce a realistic looking latent fingerprint with occluded areas of the ridge pattern.

Similarly, the ControlNet aspect of the ID-Net model will not produce realistic non-linear distortions to the generated images because it would modify the input fingerprint pattern supplied as the ControlNet input. Therefore, we also randomly sample realistic distortion grids to apply to the ControlNet image for each generation. These realistic distortion grids are obtained by computing minutiae displacements between genuine fingerprint pairs within the training dataset. During inference, an example distortion grid, indexed by the specified fingerprint acquisition type, is sampled and applied to the input reference image.

\section{Experimental Results}
In this section, we first evaluate the realism of GenPrint-generated images compared to real fingerprint images and other baseline fingerprint generation methods. We then verify the validity of each of the explainable control factors which GenPrint is trained to generate, including control over the fingerprint class, acquisition, sensor, and quality. Next, we examine GenPrint's adaptability for zero-shot style generation by using GenPrint to generate fingerprint images following the style characteristics of the unseen Latent Fingerprint in the Wild (LFIW) dataset consisting of three new latent fingerprint types, one optical sensor, one capacitive sensor, and one contactless sensor~\cite{liu2024latent}. Additionally, we evaluate the utility of GenPrint-generated images for training fingerprint recognition models and compare it with other fingerprint generation methods both when training on only synthetic images and for augmenting a set of real fingerprint images with additional synthetic fingerprint impressions. Furthermore, we show the potential for GenPrint images to be used for evaluation of fingerprint recognition models as a replacement for real fingerprint images in large-scale identification experiments. Next, we verify the uniqueness and independence of GenPrint-generated finger identities compared to the set of training fingerprint identities from which it was trained. Finally, we include a discussion on the failure cases and limitations of GenPrint.

\begin{figure}
\includegraphics[width=\linewidth]{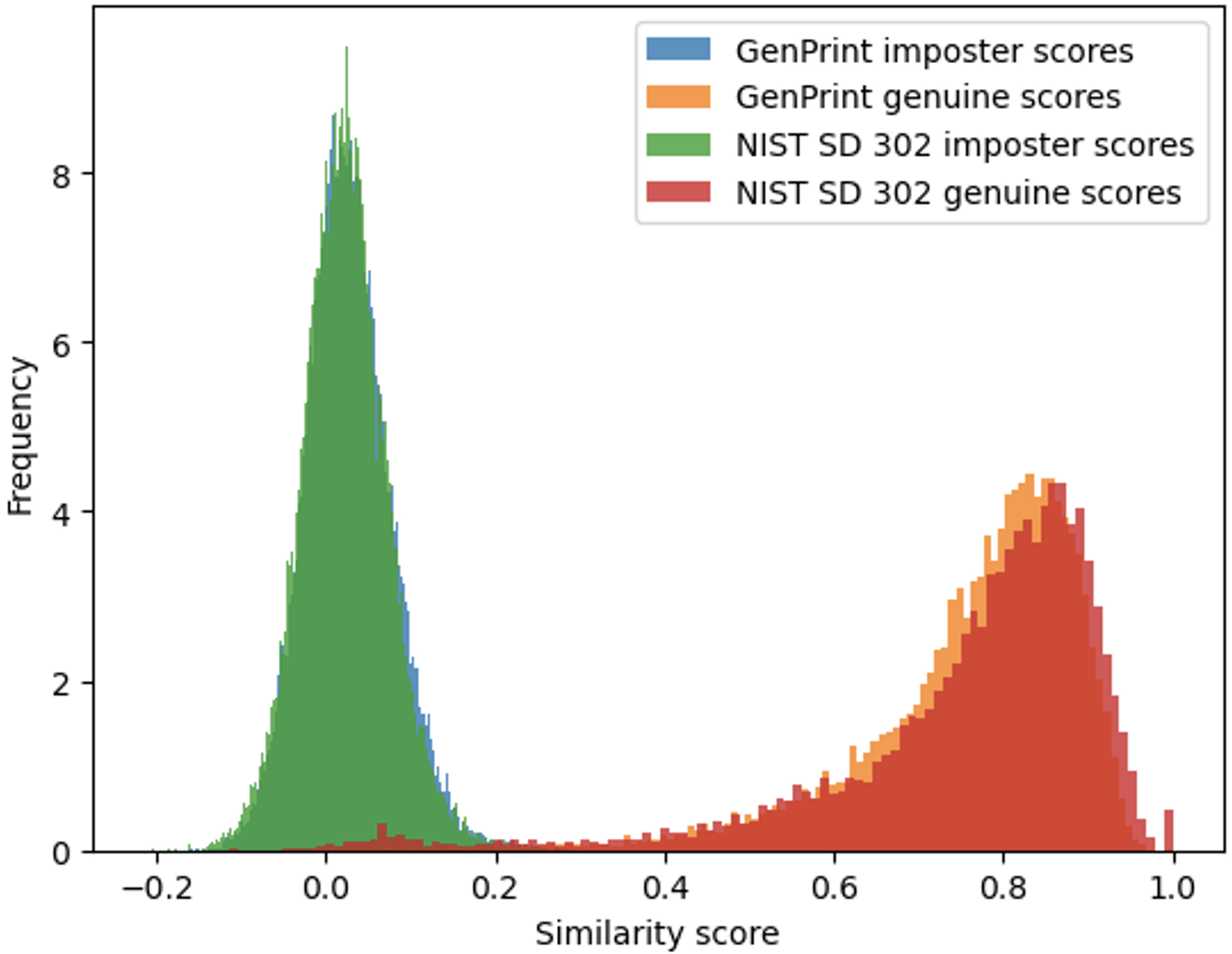} 
\caption{AFR-Net similarity score distributions for NIST SD302 real dataset and similar GenPrint dataset.}
\label{fig:pretrained_hists}
\end{figure}

\subsection{Realism of Generated Fingerprints}
\label{sec:realism}
To validate the realism of GenPrint-generated fingerprints, we performed two experiments: i.) comparing the genuine and imposter score distribution of GenPrint images to a similar composition of real fingerprint images and ii.) comparing various fingerprint and minutiae related statistics between real images and GenPrint synthetic images.

For the first experiment, we generated 400 unique synthetic fingers with 12 impressions each across a random selection of slap, rolled, and contactless fingerprints using GenPrint to mimic the size and sensor distribution of a test split of the NIST SD302 dataset consisting of 400 real finger identities with roughly 12 impressions each and a mix of different sensor and acquisition types. We then computed genuine (same identity) and imposter (different identity) score distributions using a pretrained AFR-Net~\cite{grosz2022afr} fingerprint recognition model for both the GenPrint-generated dataset and the test split of NIST SD302. The results are shown in Figure~\ref{fig:pretrained_hists}, showcasing the similarity between the score distributions of the real and generated datasets. We chose NIST SD302 for this comparison as it encompasses many of the different acquisition types (rolled, slap, contactless) that GenPrint is trained to generate. The realism of GenPrint-generated images is evident from the overlap in the distributions compared to the real fingerprint dataset. The recognition performance of each of the datasets (real and synthetic) is also very similar. For NIST SD302 and the corresponding GenPrint dataset, the true accept rate (TAR) at a false accept rate (FAR) of 0.01\% is 96.33\% and 97.33\%, respectively. 

Next, we compare various fingerprint statistics from 1,000 real, rolled fingerprint impressions from the NIST SD4 dataset to 1,000 synthetic rolled impressions generated by GenPrint and the baseline method PrintsGAN. The specific metrics being compared are summarized in Table~\ref{tab:stats} and include fingerprint area, minutiae count, and average quality of minutiae. Compared to the real fingerprint dataset, GenPrint and PrintsGAN images differ slightly between average fingerprint area compared to the real images, where PrintsGAN tends to produce smaller fingerprints and GenPrint tends to produce larger fingerprints. To normalize for the relative differences in fingerprint area, we computed the minutiae statistics on a center crop of $256\times256$ pixels. Compared to the real images, GenPrint images exhibit a higher degree of similarity than PrintsGAN images in terms of average minutiae count and quality. For example, GenPrint differs from the real fingerprint images in average minutiae count by 5.27, whereas PrintsGAN differs by 6.52. Similarly, GenPrint differs in minutiae quality by 0.02, whereas PrintsGAN differs by 1.15.

\begin{table}
\caption{Fingerprint statistics comparison of GenPrint and PrintsGAN generated images to real fingerprint images.}
\label{tab:stats}
\begin{tabular}{cccc}
\toprule
 & \begin{tabular}[c]{@{}c@{}}MSP~\cite{yoon2015longitudinal} \\ (real dataset)\end{tabular} & \begin{tabular}[c]{@{}c@{}}PrintsGAN\end{tabular} & \begin{tabular}[c]{@{}c@{}}GenPrint\end{tabular} \\
\toprule
\begin{tabular}[c]{@{}c@{}}Minutiae\\ count\end{tabular} & 37.18 $\pm$ 9.75 & 30.66 $\pm$ 6.98 & 42.45 $\pm$ 8.52 \\
\midrule
\begin{tabular}[c]{@{}c@{}}Minutiae\\ quality\end{tabular} & 80.38 $\pm$ 10.36 & 81.53 $\pm$ 9.52 & 80.40 $\pm$ 9.83 \\
\midrule
\begin{tabular}[c]{@{}c@{}}Area\\ (pixels)\end{tabular} & 192,285 $\pm$ 34,368 & 175,460 $\pm$ 25,189 & 211,599 $\pm$ 19,241 \\
\bottomrule
\end{tabular}
\end{table}

\subsection{Consistency of Control Factors}
In this section, we evaluate all the different explicit control factors which GenPrint is trained to accommodate via text prompts, including control over the fingerprint class, acquisition, sensor, and quality level. 

\begin{figure}
\includegraphics[width=\linewidth]{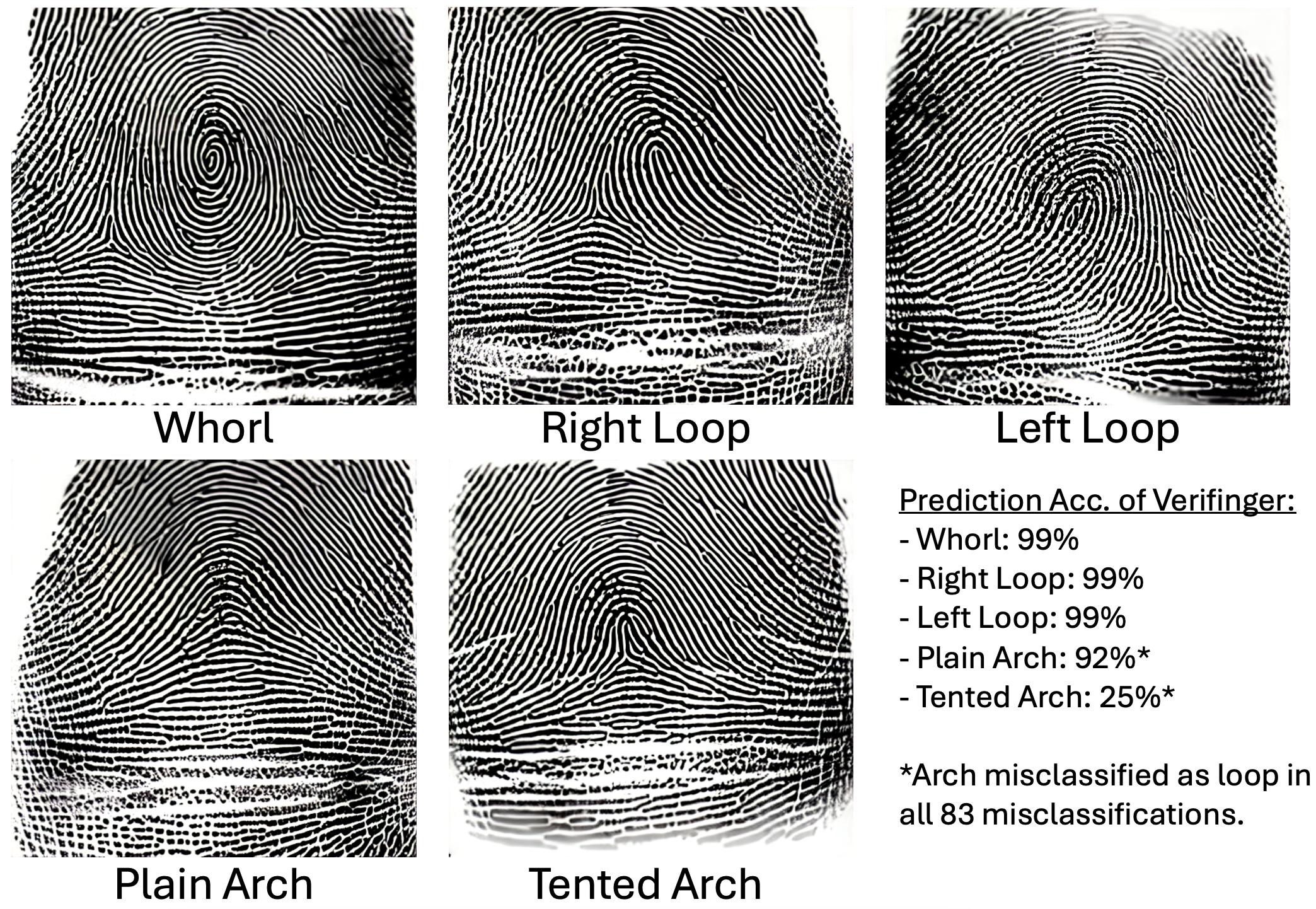} 
\caption{Example GenPrint images of different fingerprint classes and corresponding classification accuracy of Verifinger v12.4 SDK.}
\label{fig:class_acc}
\end{figure}

\subsubsection{Fingerprint Class}
GenPrint is able to generate fingerprints of any of the five major classes of fingers: whorl, left loop, right loop, plain arch, and tented arch. Examples of each of the categories generated by GenPrint are shown in Figure~\ref{fig:class_acc}. The consistency of GenPrint-generated images in following the fingerprint class prompt provided by the user is validated quantitatively using the commercially available fingerprint recognition software, Verifinger SDK v12.4. Specifically, we generate 100 unique finger identities using GenPrint in each of the five different fingerprint classes and classify each of the fingerprints using Verifinger and compute the accuracy between the Verifinger predictions and the ground truth class assigned by the input text prompts. The classification accuracy for whorl, left loop, and right loop fingerprints was 99\%, indicating that 99 out of 100 generated fingerprints were classified by Verifinger as the same class intended to be generated by GenPrint. It turned out that the classification accuracy for Verifinger on the plain arch (92\%) and tented arch types (25\%) was much more challenging for Verifinger, which often misclassified the arch type as either left or right loop in all the misclassifications. Understandably, these two fingerprint classes can be difficult to distinguish given the similarity in the ridge patterns.

\begin{figure*}
\includegraphics[width=\linewidth]{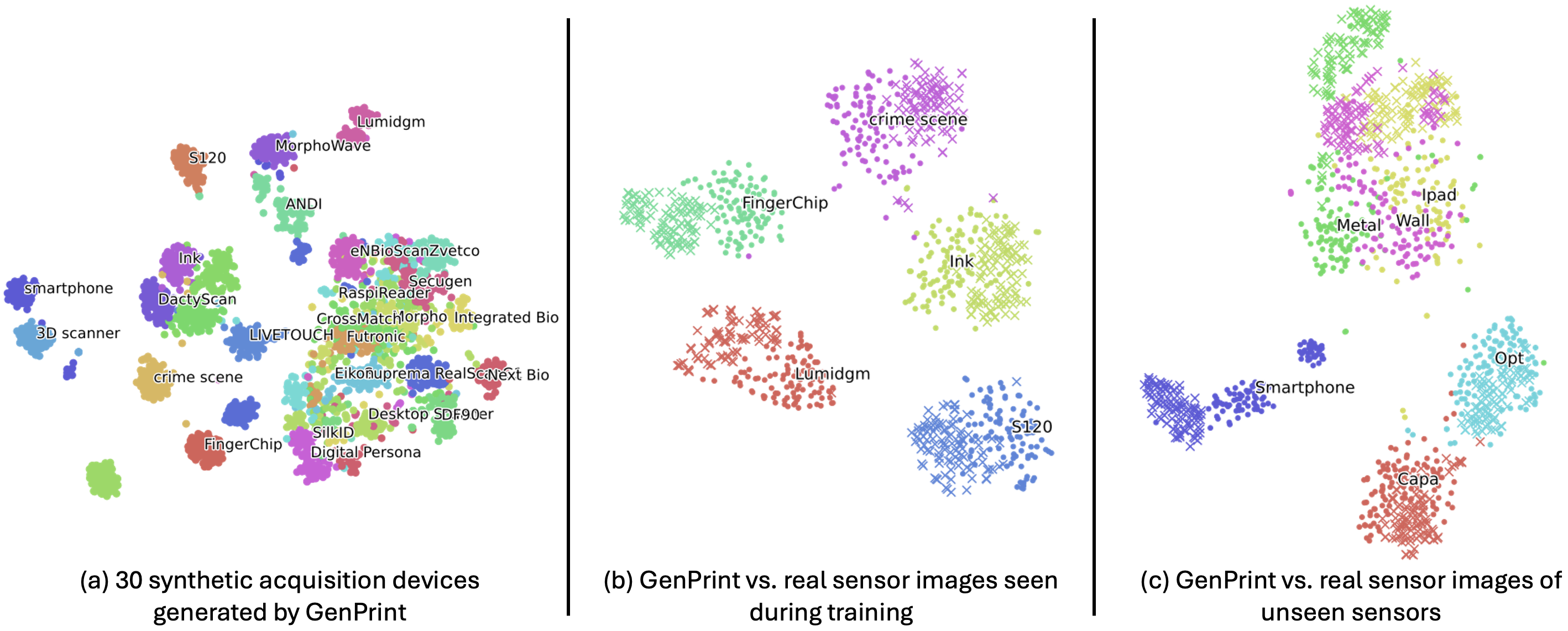} 
\caption{T-SNE plots to show (a) separation of GenPrint-generated images from different acquisition devices, (b) similarity of GenPrint images and corresponding real images of the same acquisition device, (c) similarity of zero-shot generated images to corresponding real images of novel acquisition devices which were not included in the training set of GenPrint.}
\label{fig:tsne}
\end{figure*}

\begin{figure*}
\includegraphics[width=\linewidth]{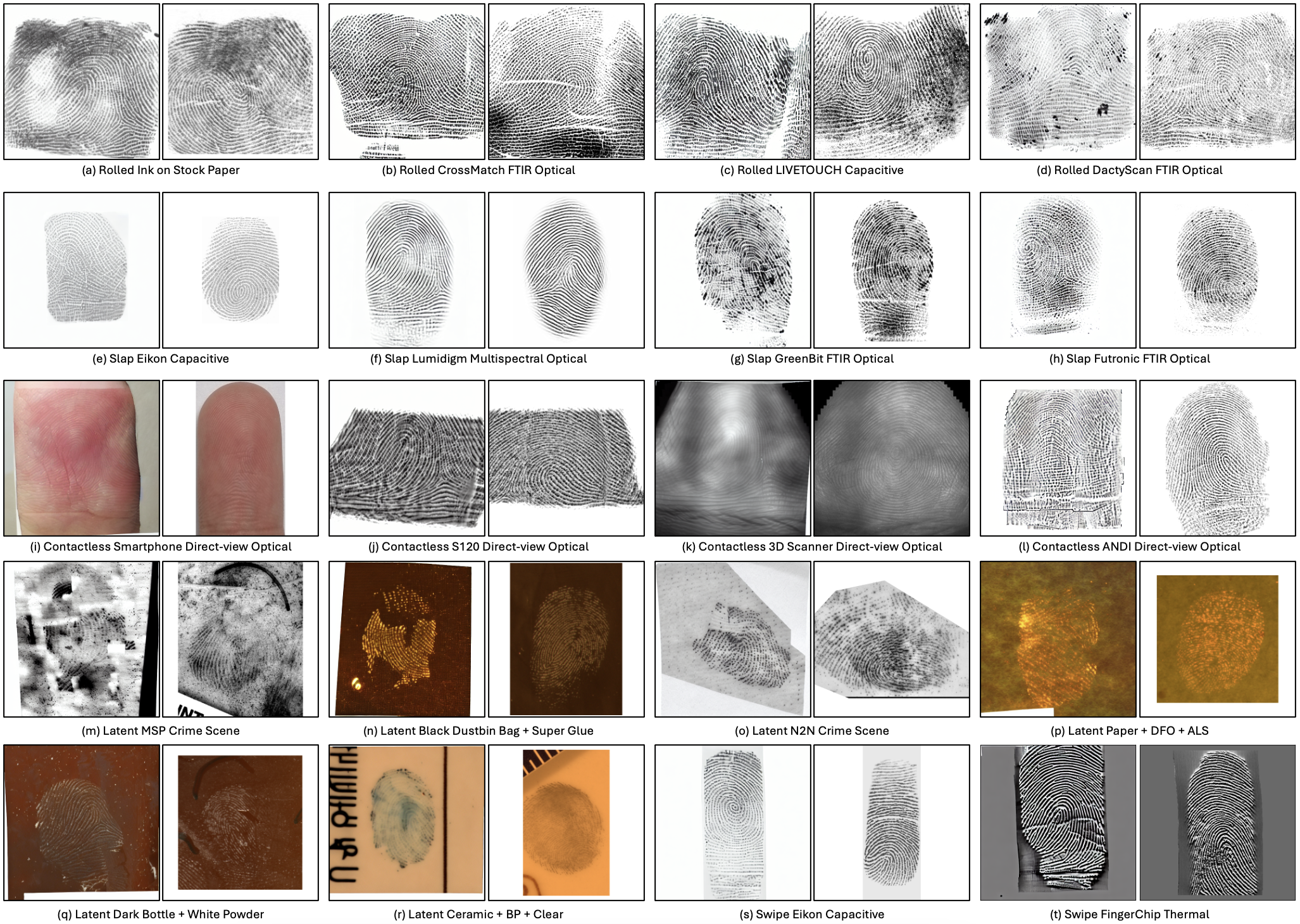} 
\caption{Example GenPrint-generated and real fingerprint images from corresponding acquisition device domains. In each pair, the left image is generated by GenPrint and the right image is a real fingerprint image of the same acquisition device to show the similarity of GenPrint images to real images with corresponding sensor characteristics.}
\label{fig:example_images}
\end{figure*}

\subsubsection{Fingerprint Acquisition and Sensor Type}
GenPrint is trained on data from 30 different acquisition devices which consist of various rolled, slap, swipe, contactless, and latent fingerprint acquisition types. Some example images from different devices are given in Figure~\ref{fig:example_images} along with corresponding GenPrint-generated images in those same device domains, where the left image in each pair is a synthetic fingerprint generated by GenPrint, and the right image is an example fingerprint image from a real fingerprint database. Comparing GenPrint images and corresponding real images in the same sensor and acquisition types highlights the realism and diversity in the possible generation space of GenPrint.

To visualize the separability of all 30 acquisition device characteristics that GenPrint is trained to generate, we first generated 100 example fingerprint images in each acquisition device domain. Then, we extracted representation embeddings using a pretrained VGG network and plotted them in t-SNE~\cite{van2008visualizing} embedding space. The result is shown in subfigure (a) of Figure~\ref{fig:tsne} which shows clear separation between very distinct acquisition devices and some small overlap in similar sensors, such as the large number of different slap FTIR optical devices sharing similar characteristics. Furthermore, we also generated VGG embeddings for 100 real fingerprint image examples in 5 different acquisition devices and embedded them into the t-SNE space along with their corresponding generated images from GenPrint to show the similarity between corresponding real and synthetic images of the same acquisition device domains.

\subsubsection{Quality Control}
\begin{figure}
\includegraphics[width=\linewidth]{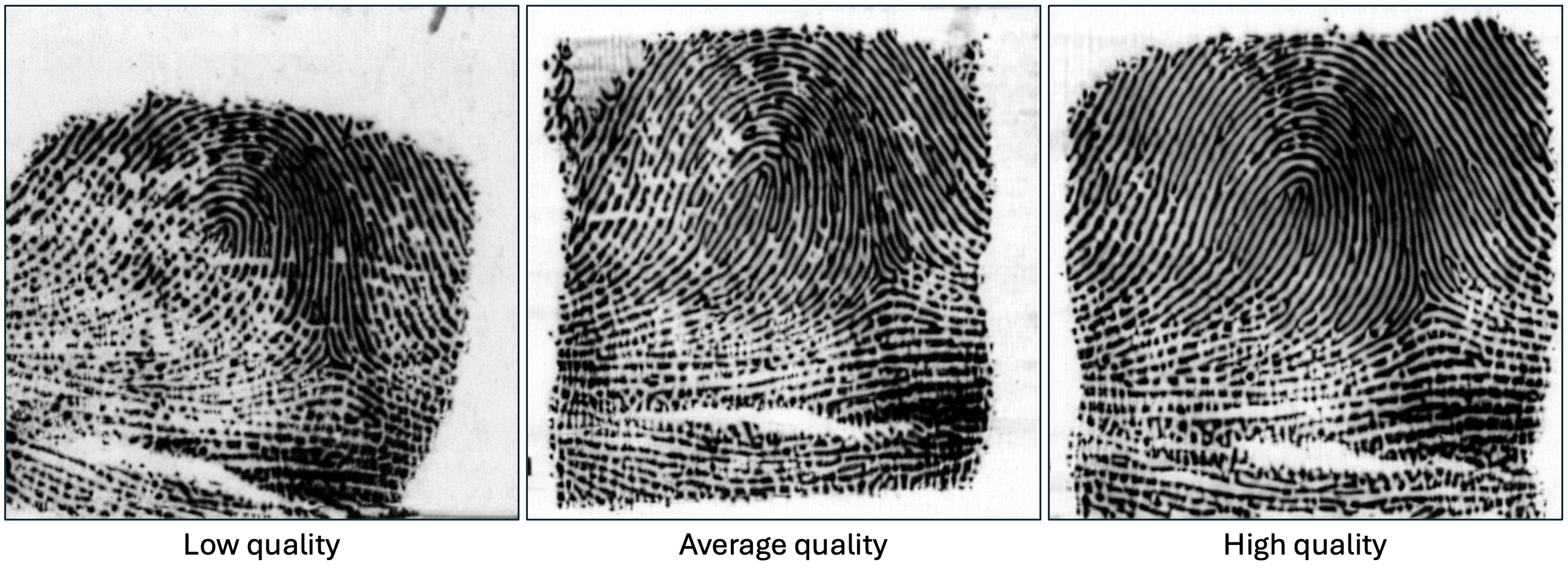} 
\caption{Example low, average, and high quality rolled fingerprint impressions of a finger generated by GenPrint.}
\label{fig:example_quality_images}
\end{figure}

\begin{figure*}
\includegraphics[width=\linewidth]{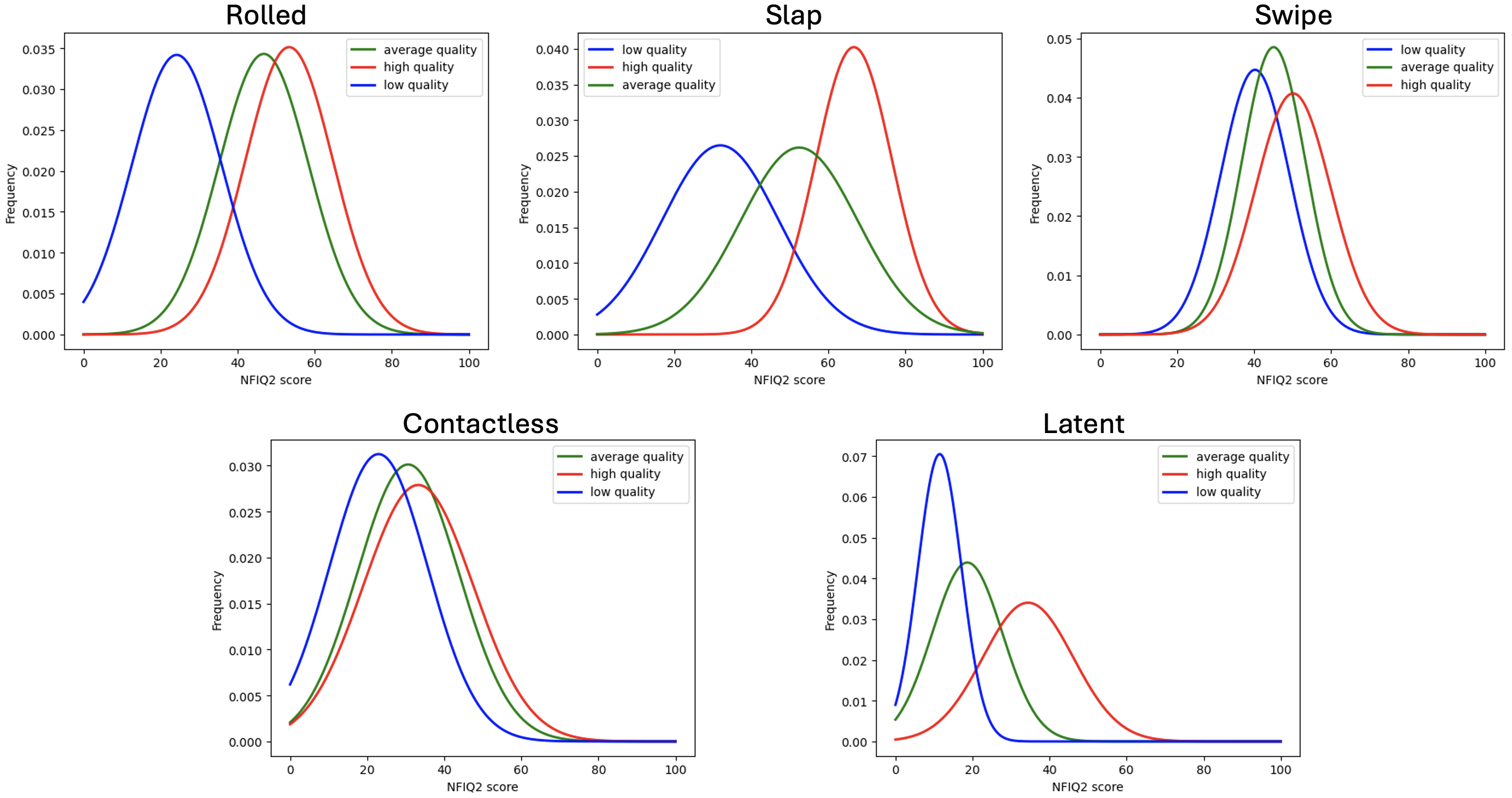} 
\caption{NFIQ 2.0 score distributions for fingerprints generated by GenPrint across five different fingerprint acquisition types.}
\label{fig:nfiq}
\end{figure*}

There are two ways in which GenPrint can manipulate the quality of the generated images. The first is through the text prompt where the user can specify either low, average, or high quality, and the other is through passing a reference style image with a relatively low, average, or high quality appearance. Empirically, we found both approaches to work well. For validating the quality control of GenPrint, we generated datasets of 100 unique synthetic finger identities with 300 impressions of each of the five different acquisition types (rolled, slap, swipe, contactless, and latent) and used the text prompt to generate 100 of those impressions for each quality level (low, average, and high). Example low, average, and high quality images for three generated rolled fingerprints are given in Figure~\ref{fig:example_quality_images} for visualization. We then computed the NFIQ 2.0 quality score using the Verifinger SDK and plotted the quality distributions in Figure~\ref{fig:nfiq}. There is clear separation among each of the quality levels across each of the acquisition types, verifying GenPrint's appropriate control over the quality of the generated fingerprints.

\subsection{Zero-shot Fingerprint Style Generation}
\label{sec:zeroshot}
To validate the quality of zero-shot fingerprint style generation, we performed one last experiment using t-SNE visualizations where we embed 100 example synthetic and real images from 6 different acquisition device domains from an unseen dataset which was not included in the training dataset for GenPrint. These images come from the recently released LFIW dataset~\cite{liu2024latent}. Again, we observe very close similarity to corresponding real and synthetic images of the same acquisition devices, demonstrating GenPrint's adaptability toward zero-shot style generation from novel acquisition devices.

\subsection{Utility for Training Fingerprint Recognition Models}
One of the most important criteria for the quality of synthetic fingerprint generators is their utility for training fingerprint recognition models. We evaluate the utility of GenPrint both when training on only synthetically generated images and when augmenting a set of real fingerprint images with additional synthetic data. We compare with several previous synthetic fingerprint generators as baselines including SFinGe, PrintsGAN, and FPGAN-Control.

\begin{figure*}
\includegraphics[width=\textwidth]{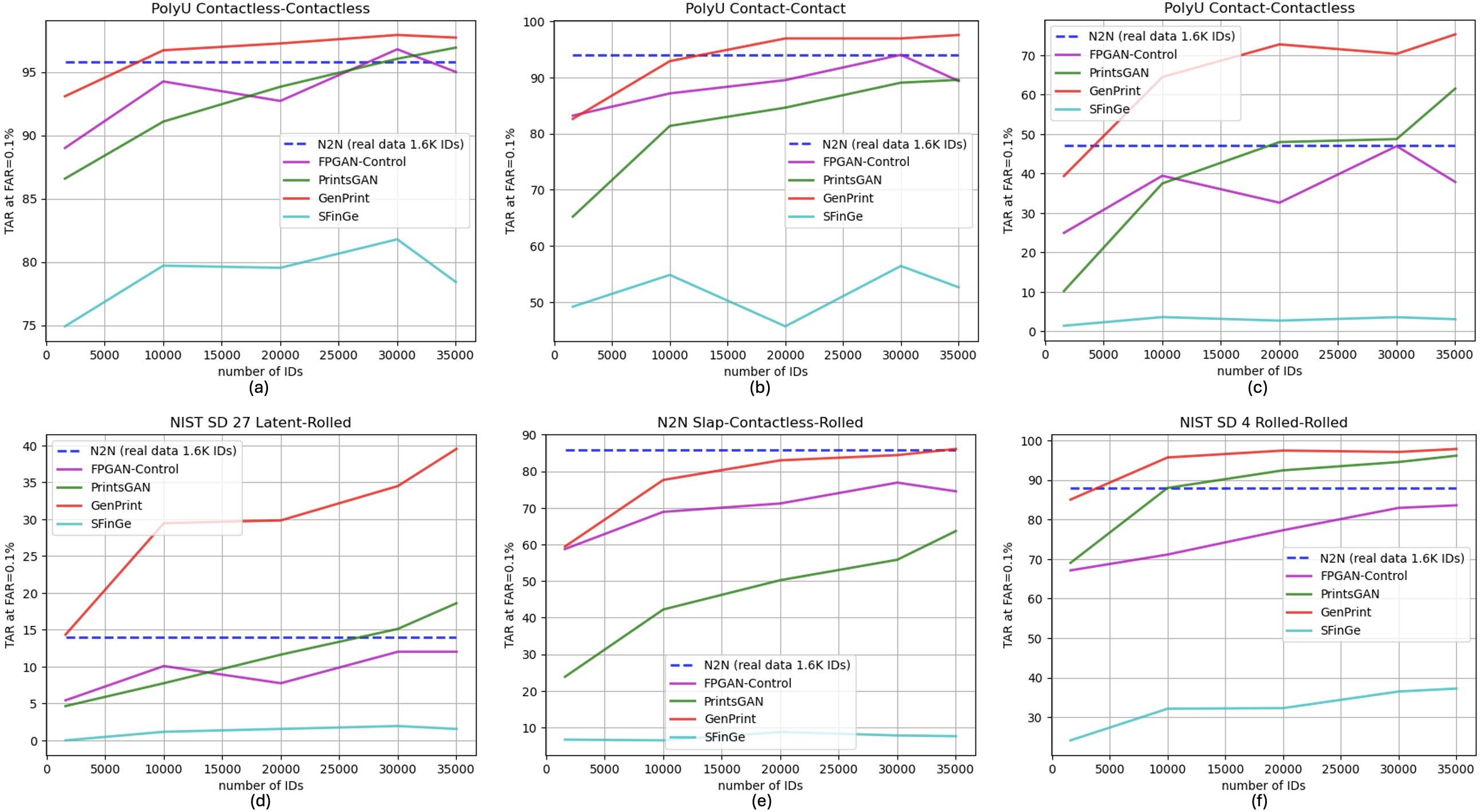} 
\caption{Authentication accuracy (TAR at FAR=0.1\%) of ResNet50 trained on synthetic data from various fingerprint generation methods including the proposed GenPrint.}
\label{fig:synth_comparison}
\end{figure*}

For the first experiment, we generate synthetic databases of 35,000 identities and 15 impressions per identity using each synthetic generation method. Some example images from each method are shown in Figure~\ref{fig:baseline_imgs}. We then train ResNet50~\cite{he2016deep} recognition models using an ArcFace loss function on incremental subsets of each database using increments from 1,600 identities (the size of the real N2N fingerprint database) to 35,000 identities and plot the performance of the trained models on various evaluation datasets (see Figure~\ref{fig:synth_comparison}). The evaluation datasets used are summarized in Table~\ref{tab:test_datasets} and include fingerprint impressions of diverse acquisition devices including rolled, slap, contactless, and latent fingerprint types. We also summarize the TAR at an FAR of 0.1\% for training on 35,000 identities from each method in Table~\ref{tab:aug_comparison}. From Figure~\ref{fig:synth_comparison}, we can clearly see that the performance of the recognition model trained on GenPrint images performs far better than any of the baseline synthetic methods and even surpasses the performance of training on the real N2N fingerprint dataset as the number of synthetic identities is increased.

\begin{figure*}
\includegraphics[width=\textwidth]{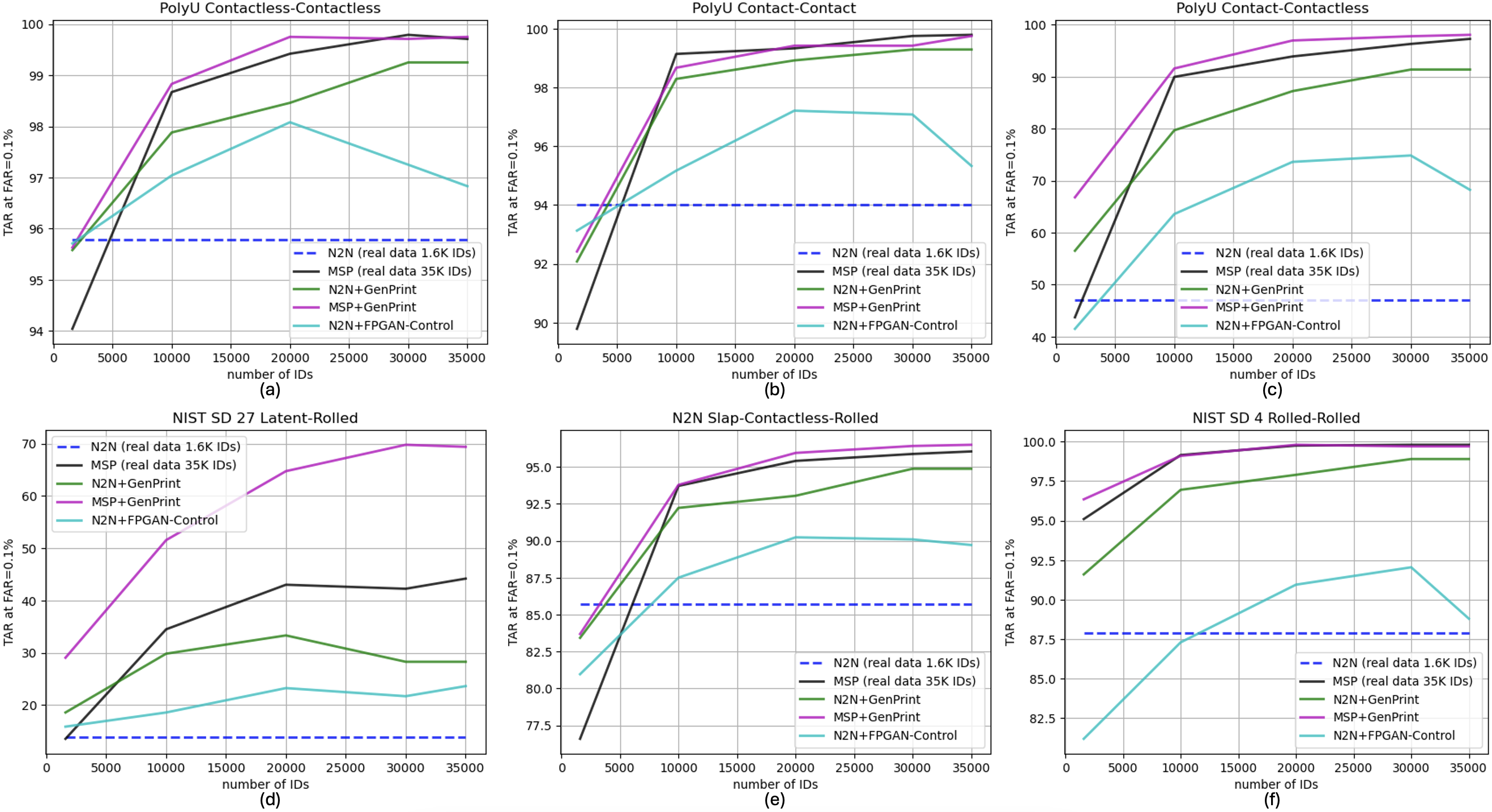} 
\caption{Authentication accuracy (TAR at FAR=0.1\%) of ResNet50 trained on a combination of real and synthetic data from FPGAN-Control and the proposed GenPrint evaluated on six different test scenarios.}
\label{fig:aug_comparison}
\end{figure*}

For the second experiment, we compared the utility of GenPrint to the next best performing synthetic method FPGAN-Control in augmenting an existing set of real fingerprint data for training on a combination of real and synthetic. Starting from the initial set of 1,600 real finger identities from N2N, we add increasing amounts of synthetic identities and again plot the performance of the trained ResNet50 models as the number of identities is increased. The results in Figure~\ref{fig:aug_comparison} show that both synthetic methods improve the performance when used for augmentation, but the improvement from GenPrint images is far superior.

The previous experiment showcased improvement of augmenting a limited set of real fingerprint data of only 1,600 unique finger identities, but naturally a question arises as to whether synthetic data augmentation is still helpful if the number of unique, real fingerprint identities in the training set is already large (e.g., 35,000). To investigate this question given that the number of identities is already large, rather than include additional synthetic identities, we instead take the existing real identities and use GenPrint to synthesize additional impressions in a more diverse range of acquisition devices. For this experiment, we use 35,000 unique fingerprint identities from the Michigan State Police (MSP) longitudinal fingerprint dataset~\cite{yoon2015longitudinal} which has about 12 impressions per identity and augment each finger identity with an additonal 15 synthetic impressions of various acquisition devices. The result of augmenting MSP with GenPrint impressions is shown in Figure~\ref{fig:aug_comparison}. As the number of identities increases, the plots show that GenPrint does indeed improve the performance significantly by augmenting the diversity of the already existing fingerprint images. This improvement is particularly evident when the test datasets contain sensor characteristics not included in the original MSP dataset but which GenPrint is able to synthesize (e.g., contactless and latent fingerprints). For reference, the TAR at FAR=0.1\% using 35,000 training identities is given in Table~\ref{tab:aug_comparison}. 

In both of the previous experiments, we trained only ResNet50 models for the comparison. Thus, we now study the impact of additional model architectures and examine whether similar trends arise. In particular, we train two additional model architectures on both GenPrint and FPGAN-Control datasets of 35,000 identities and 15 impressions per identity. These include a ResNet18 model and a vision transformer (ViT)~\cite{dosovitskiy2020image} with a patch size of 16 and 12 layers. As shown in Table~\ref{tab:architectures}, the same relative performance gap between training on GenPrint vs. FPGAN-Control images across each model architecture is consistent with our more extensive experiments using ResNet50.

\begin{table*}
\caption{Test Datasets.}
\label{tab:test_datasets}
\rowstretch{1.5} 
\begin{tabularx}{\textwidth}{>{\centering\arraybackslash}p{0.18\linewidth}|>{\centering\arraybackslash}p{0.12\linewidth}|>{\centering\arraybackslash}p{0.45\linewidth}|>{\centering\arraybackslash}p{0.15\linewidth}}
\toprule[1pt]
\textbf{Test Dataset} & \textbf{Acquisition Types} & \textbf{Sensor Types} & \textbf{No. Images (Fingers)} \\
\Xhline{1pt}
PolyU Contactless 2D to Contact-based 2D~\cite{lin2018cnn} & Slap, Contactless & Digital Persona, smartphone & 1920 (336) \\
\hline
NIST SD 4~\cite{sd4} & Rolled & Ink on paper & 4000 (2,000) \\
\hline
NIST SD 302~\cite{nist302} & Slap, Contactless, Rolled, & CrossMatch, Eikon, GreenBit, ANDI, S120, MorphoWave, DactyScan, LIVETOUCH, Futronic, RaspiReader & 2548 (400) \\
\hline
NIST SD 27~\cite{sd27} & Rolled, Latent & Ink on paper, crime scene & 1032 (258) \\
\bottomrule[1pt]
\end{tabularx}
\end{table*}

\begin{table*}
\centering
\caption{Authentication accuracy (TAR at FAR=0.1\%) of ResNet50 trained on synthetic data from various fingerprint generators including the proposed GenPrint. A ResNet50 model trained on N2N, a real dataset, is included as a baseline.}
\label{tab:synth_comparison}
\resizebox{\textwidth}{!}{%
\begin{tabular}{lcccccccc}
\toprule
Training Data & No. IDs & No. imgs/ID & \begin{tabular}[c]{@{}c@{}}N2N\\ slap-rolled-contactless\end{tabular} & \begin{tabular}[c]{@{}c@{}}NIST SD4\\ rolled-rolled\end{tabular} & \begin{tabular}[c]{@{}c@{}}PolyU\\ contact-contact\end{tabular} & \begin{tabular}[c]{@{}c@{}}PolyU\\ contactless-contactless\end{tabular} & \begin{tabular}[c]{@{}c@{}}PolyU\\ contact-contactless\end{tabular} & \begin{tabular}[c]{@{}c@{}}NIST SD27\\ latent-rolled\end{tabular} \\
\toprule
N2N~\cite{nist302} (real dataset) & 1,600 & 12 & 85.73 & 87.90 & 94.00 & 95.79 & 47.08 & 13.95 \\
\midrule
SFinGe~\cite{cappelli2004sfinge} & 35,000 & 15 & 7.62 & 37.25 & 52.63 & 78.42 & 3.07 & 1.55 \\
\midrule
FPGAN-Control~\cite{shoshan2024fpgan} & 35,000 & 15 & 74.52 & 83.60 & 89.38 & 95.00 & 37.86 & 12.02 \\
\midrule
PrintsGAN~\cite{engelsma2022printsgan} & 35,000 & 15 & 63.66 & 96.15 & 89.58 & 96.92 & 61.51 & 18.60 \\
\midrule
GenPrint & 35,000 & 15 & \textbf{86.08} & \textbf{97.85} & \textbf{97.58} & \textbf{97.71} & \textbf{75.26} & \textbf{39.53}\\
\bottomrule
\end{tabular}%
}
\end{table*}

\begin{table*}
\centering
\caption{Authentication accuracy (TAR at FAR=0.1\%) of ResNet50 trained on a combination of real and synthetic data from FPGAN-Control and the proposed GenPrint evaluated on six different test scenarios.}
\label{tab:aug_comparison}
\resizebox{\textwidth}{!}{%
\begin{tabular}{lcccccccc}
\toprule
Training Data & No. IDs & No. images/ID & \begin{tabular}[c]{@{}c@{}}N2N\\ slap-rolled-contactless\end{tabular} & \begin{tabular}[c]{@{}c@{}}NIST SD4\\ rolled-rolled\end{tabular} & \begin{tabular}[c]{@{}c@{}}PolyU\\ contact-contact\end{tabular} & \begin{tabular}[c]{@{}c@{}}PolyU\\ contactless-contactless\end{tabular} & \begin{tabular}[c]{@{}c@{}}PolyU\\ contact-contactless\end{tabular} & \begin{tabular}[c]{@{}c@{}}NIST SD27\\ latent-rolled\end{tabular} \\
\toprule
N2N~\cite{nist302} (real dataset) & 1,600 & 12 & 85.73 & 87.90 & 94.00 & 95.79 & 47.08 & 13.95 \\
\midrule
N2N~\cite{nist302} + FPGAN-Control~\cite{shoshan2024fpgan} & 35,000 & 15 & 89.71 & 88.80 & 95.33 & 96.83 & 68.25 & 23.64 \\
\midrule
N2N~\cite{nist302} + GenPrint & 35,000 & 13.5 & 94.69 & 98.90 & 99.54 & 99.17 & 90.90 & 46.51 \\
\midrule
MSP~\cite{yoon2015longitudinal} (real dataset) & 35,000 & 12 & 96.04 & \textbf{99.80} & \textbf{99.79} & 99.71 & 97.29 & 62.02 \\
\midrule
MSP~\cite{yoon2015longitudinal} + GenPrint & 35,000 & 27 & \textbf{96.49} & 99.70 & 99.75 & \textbf{99.75} & \textbf{98.07} & \textbf{69.38}\\
\bottomrule
\end{tabular}%
}
\end{table*}

\begin{table*}
\centering
\caption{Training on 35K IDs, 15 impressions with different model architectures and evaluated on six different test scenarios. Results reported as TAR @ FAR=0.1\%.}
\label{tab:architectures}
\resizebox{\textwidth}{!}{%
\begin{tabular}{cccccccc}
\toprule
Model & Training Dataset & \begin{tabular}[c]{@{}c@{}}NIST SD302\\ slap-rolled-contactless\end{tabular} & \begin{tabular}[c]{@{}c@{}}NIST SD4\\ rolled-rolled\end{tabular} & \begin{tabular}[c]{@{}c@{}}PolyU\\ contact-contact\end{tabular} & \begin{tabular}[c]{@{}c@{}}PolyU\\ contactless-contactless\end{tabular} & \begin{tabular}[c]{@{}c@{}}PolyU\\ contact-contactless\end{tabular} & \begin{tabular}[c]{@{}c@{}}NIST SD27\\ latent-rolled\end{tabular} \\
\toprule
ResNet18 & FPGAN-Control~\cite{shoshan2024fpgan} & 61.92 & 70.60 & 80.79 & 87.83 & 26.70 & 8.14\\
\midrule
ResNet18 & GenPrint & \textbf{71.84} & \textbf{91.75} & \textbf{90.25} & \textbf{93.04} & \textbf{42.45} & \textbf{23.64} \\
\midrule[1pt]
ResNet50 & FPGAN-Control~\cite{shoshan2024fpgan} & 74.52 & 83.60 & 89.38 & 95.00 & 37.86 & 12.02\\
\midrule
ResNet50 & GenPrint & \textbf{86.08} & \textbf{97.85} & \textbf{97.58} & \textbf{97.71} & \textbf{75.26} & \textbf{39.53} \\
\midrule[1pt]
ViT & FPGAN-Control~\cite{shoshan2024fpgan} & 45.47 & 70.40 & 45.92 & 83.33 & 6.28 & 5.43\\
\midrule
ViT & GenPrint & \textbf{79.81} & \textbf{96.80} & \textbf{95.75} & \textbf{96.71} & \textbf{61.25} & \textbf{30.62}\\
\bottomrule[1pt]
\end{tabular}%
}
\end{table*}

\subsection{Utility for Evaluating Fingerprint Recognition Models}
\begin{figure}
\includegraphics[width=\columnwidth]{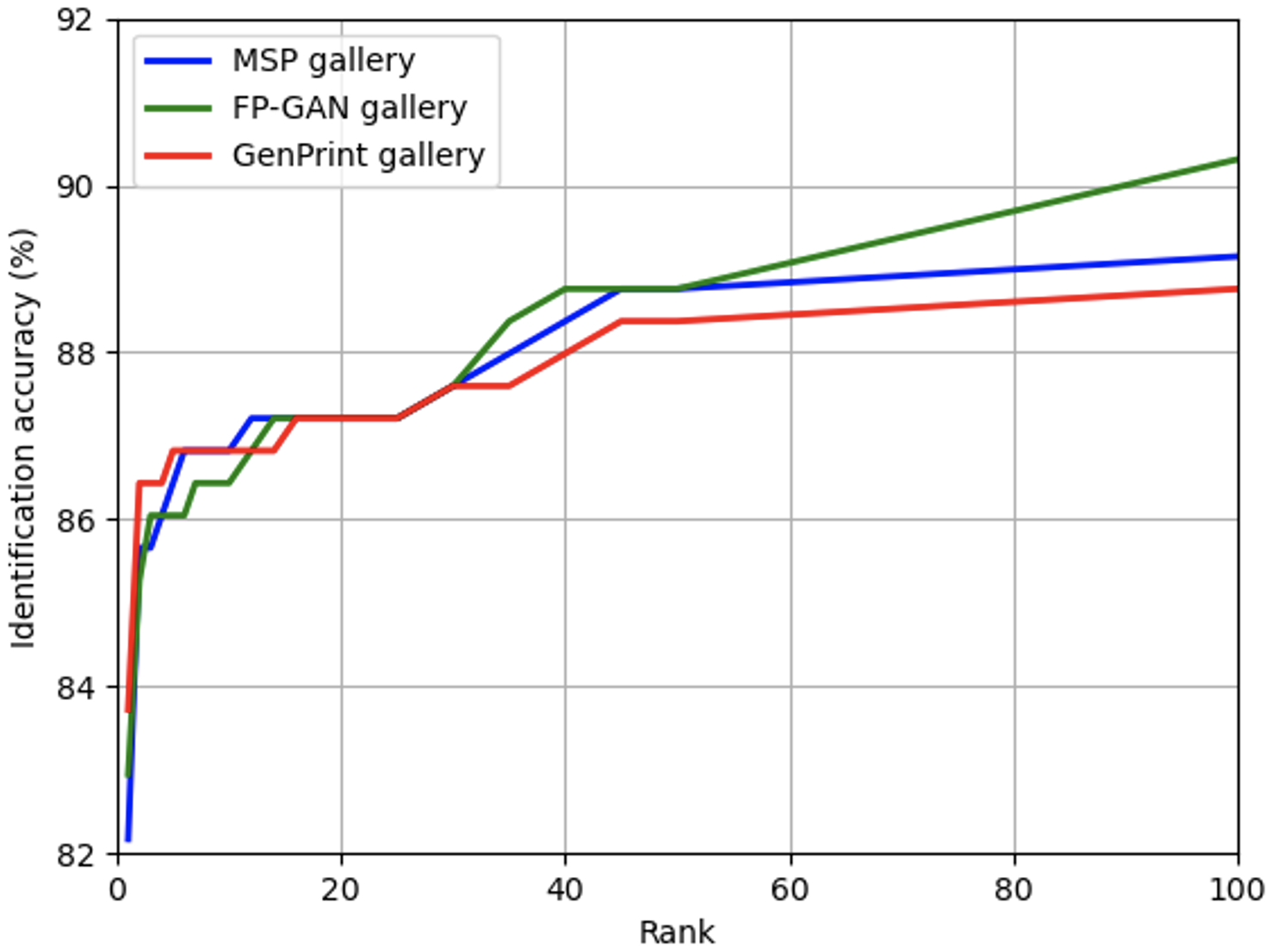} 
\caption{Search results using probes from NIST SD 27 and 64,000 identity background from GenPrint compared to the real MSP dataset and FPGAN-Control backgrounds.}
\label{fig:search}
\end{figure}

In addition to being useful for training, synthetic fingerprints can also help with large-scale evaluation of fingerprint recognition algorithms, where collecting a dataset of potentially millions of unique real fingers can be prohibitively expensive. To demonstrate the feasibility of GenPrint images to be used for such purposes, we generated a large database of 64,000 unique rolled fingerprints to compare with a database of 64,000 real rolled fingerprint identities from the MSP dataset as a background gallery for latent to rolled fingerprint search using latent probes and corresponding mates from the NIST SD 27 latent dataset. Ideally, the search performance should be similar when using the real fingerprint background images and GenPrint fingerprint background images. We repeated the experiment using a database of 64,000 unique identities from FPGAN-Control as a baseline. The results on the three different gallery backgrounds are given in Figure~\ref{fig:search}, which shows better overlap in the search accuracies between GenPrint background gallery and the real fingerprint gallery compared to the overlap between FPGAN-Control and the real dataset, indicating that GenPrint images make a more suitable replacement for real images for large-scale search evaluations than the baseline FPGAN-Control method. In particular, the rank-1 accuracy on the real background dataset is 82.17\%, whereas it was 82.95\% and 83.72\% for GenPrint and FPGAN-Control, respectively.

\subsection{Biometric Capacity}
\label{sec:capacity}
Ideally, every synthetically finger identity should be unique, but the probability of encountering ``duplicate" identities, those which have a high similarity to each other, increases as the size of the dataset grows, which is true even for real fingerprint datasets. Nonetheless, the possible number of unique finger identities that a model can generate, referred to as the biometric capacity~\cite{boddeti2023biometric}, is an important factor for comparison among synthetic biometric generators. Unfortunately, accurately measuring the biometric capacity is a difficult and open question and empirically computing the similarity between all generated identities scales in complexity as O($n^2$), quickly becoming computationally demanding for anything above 100,000 identities.

One recent method by Bodetti et al.~\cite{boddeti2023biometric} proposed a geometrical model of capacity by embedding face images into a hyperspherical representation space and using a specified FAR to measure the overlap between class representations. However, this approach only aims to estimate an upper bound on the capacity and when we applied the code to our generated images and several other baselines, we received capacity estimates on the order of $10^{32}$. Instead, to obtain more practical insights, we used a pretrained AFR-Net fingerprint matcher to compute the percentage of ``duplicate" identities being generated as the number of generated identities increases for both PrintsGAN and GenPrint. We obtain duplicate identities by computing all possible imposter score comparisons between the generated identities and determine how many of the pairs produced similarity scores to each other which fell above the genuine match threshold of 0.35 computed on the real NIST SD4 dataset. This capacity comparison is fair since PrintsGAN and GenPrint were trained on fingerprint databases of similar number of identities (38,291 for PrintsGAN and 37,351 for GenPrint); however, GenPrint is based off diffusion models which are believed to better capture the full data distribution compared to GANs~\cite{dhariwal2021diffusion}. As shown in Table~\ref{tab:capacity}, the number of duplicate identities is increasing at a large rate for PrintsGAN as the number of generated identities increases, unlike GenPrint which closely follows the trend on the real MSP fingerprint dataset as the number of identities approaches 100,000.

\subsection{Identity Leakage}
\label{sec:leakage}
Besides the capacity of the biometric generator, a privacy preserving model should also not leak sensitive information from the dataset on which it was trained. In other words, the generated finger identities from GenPrint should not have high similarity with any of the finger identities in the training set. We aimed to measure the potential identity leakage of GenPrint by generating 35,000 unique synthetic fingerprint identities and computed similarity scores to each of the 37,351 real training finger identities in our training dataset using a pretrained AFR-Net fingerprint recognition model~\cite{grosz2022afr}. Out of these 35,000 synthetic identities, only 10 (0.03\%) had a similarity score with any training identity above 0.231, the genuine match threshold computed on FVC 2002 DB1A at FAR=0.01\%. Furthermore, even out of those ten similarity scores that fell above the threshold, the maximum similarity score obtained was just 0.297, only slightly above the threshold.

\begin{table}
\centering
\caption{Percentage of duplicate identities generated by PrintsGAN and GenPrint as the number of generated identities increases from 20,000 to 100,000. A duplicate identity is counted whenever an imposter score between any of the generated identities is above a genuine match threshold of 0.35, which was computed on the real NIST SD4 dataset using a pretrained AFR-Net fingerprint recognition model.}
\label{tab:capacity}
\resizebox{\linewidth}{!}{%
\begin{tabular}{cccc}
\toprule
Number of IDs & \begin{tabular}[c]{@{}c@{}}MSP\\ (real data)\end{tabular} & GenPrint & PrintsGAN~\cite{engelsma2022printsgan} \\
\toprule
20,000 & 0.060\% & 0.070\% & 4.255\% \\
40,000 & 0.078\% & 0.190\% & 7.408\% \\
60,000 & 0.103\% & 0.305\% & 9.885\% \\
80,000 & 0.154\% & 0.461\% & 12.23\% \\
100,000 & 0.170\% & 0.535\% & 14.43\% \\
\bottomrule
\end{tabular}%
}
\end{table}


\begin{figure}
\includegraphics[width=\columnwidth]{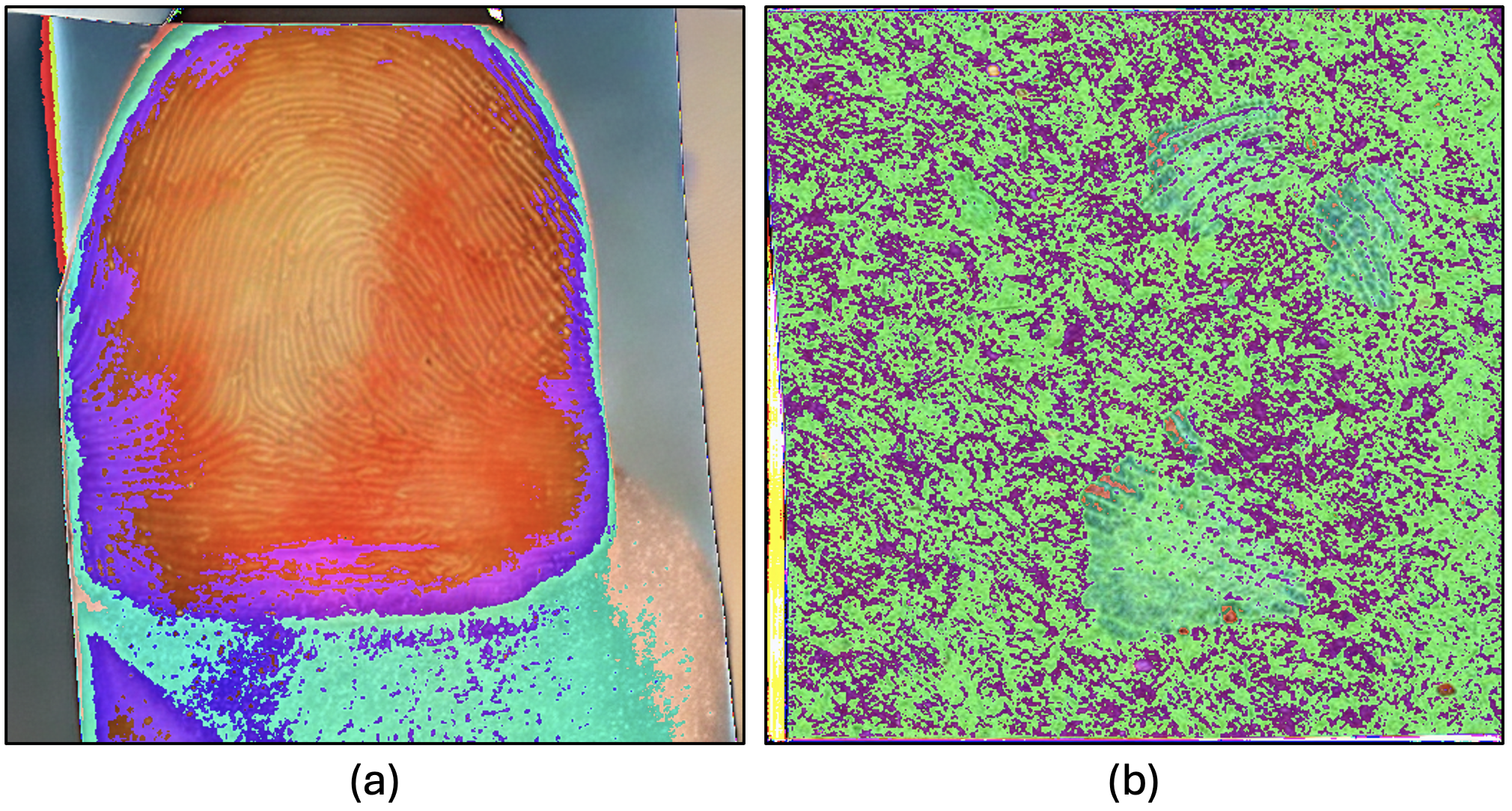} 
\caption{Example failure cases generated by GenPrint exhibiting  noise and color artifacts.}
\label{fig:failures}
\end{figure}

\begin{figure}
\includegraphics[width=\columnwidth]{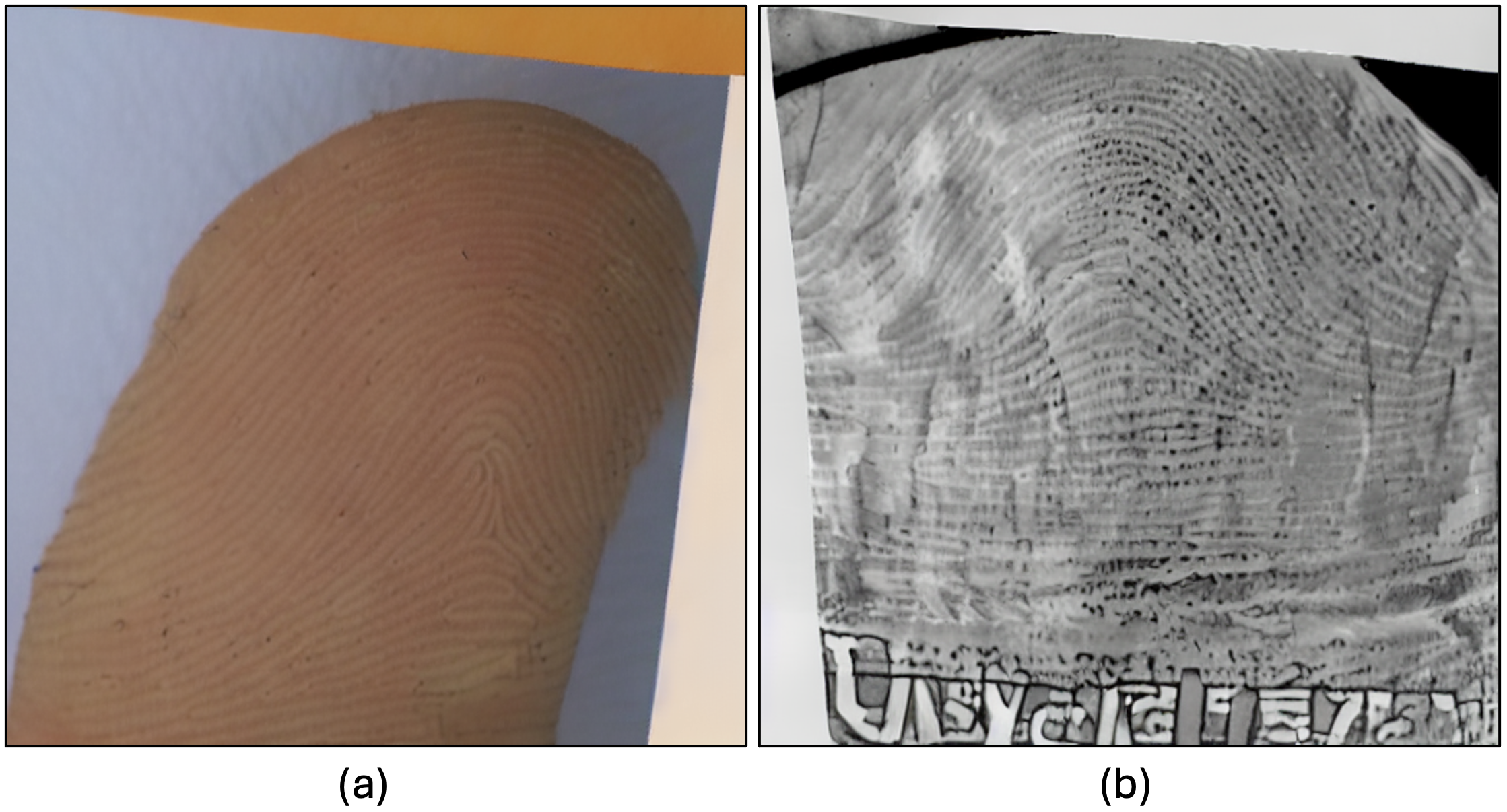} 
\caption{Example images with mixed text prompts. The image in (a) was created with a prompt containing the acquisition type of ``latent" and sensor type of ``smartphone", whereas (b) was prompted with acquisition type ``contactless" and sensor type ``crime scene".}
\label{fig:mixed}
\end{figure}

\subsection{Failures and Limitations}
On occasion, the outputs generated from GenPrint can exhibit some noise and other color artifacts. Some of these failure cases are visualized in Figure~\ref{fig:failures}. For the image in subfigure (a), the prompt was for a low quality contactless fingerprint from a smartphone camera, and the prompt used in (b) was for a low quality latent fingerprint from a crime scene. Empirically, we found that the probability of such artifacts occurring is higher when the quality of the fingerprint is prompted as low. Another potential area for unexpected outputs is in mixing acquisition and sensor types that may not be realistic. For example, the image produced in subfigure (a) of Figure~\ref{fig:mixed} was prompted with the acquisition type of ``latent" and sensor type of ``smartphone", whereas subfigure (b) was prompted with acquisition type ``contactless" and sensor type ``crime scene". These mixed prompts produce fingerprint images that resemble characteristics of both contactless and latent fingerprints but are not quite realistic. Nonetheless, even though mixing various acquisition and sensor types may produce unrealistic fingerprint images, the results may still prove to be a useful data augmentation tool for training fingerprint recognition models. In fact, all experiments conducted in this paper are without editing or removing any images generated by GenPrint.

One of the most significant limitations of GenPrint and DDPM models in general is the computational efficiency, both in terms of training and inference time and memory footprint. Training efficiency of GenPrint was partially mitigated by utilizing LoRA weights for training; however, the inference speed for GenPrint is still about 1.13 seconds per image (512$\times$512 resolution) using an Nvidia A100 GPU and an AMD EPYC 7543 32-Core Processor - which is much slower compared to GANs (e.g., 7.41 ms per image for FPGAN-Control on the same hardware). For offline generation of synthetic datasets, the latency and RAM usage are both just a nuisance; however, they prevent the model from being useful for online generation of synthetic data during training of recognition models.

\section{Conclusion}
By employing latent diffusion models with multimodal conditions, GenPrint offers a versatile framework capable of generating diverse fingerprint images while preserving identity and providing explainable control over various appearance factors. Unlike previous approaches, GenPrint is not constrained by the characteristics of the training dataset alone, allowing for the generation of novel sensor and style attributes during inference without the need for additional fine-tuning. The experimental results showcase the efficacy of GenPrint in terms of identity preservation and narrowing the gap between synthetic and real domains. Moreover, the universality of GenPrint-generated images improves model training by augmenting the diversity of existing fingerprint datasets, thus enhancing the performance and generalization of fingerprint recognition systems. The same or similar model architecture can also be applied to other areas of biometrics (e.g., face, palmprint, iris, etc.) which we are currently undertaking.

\section{Acknowledgment}
This research was supported by a grant from the Department of Homeland Security via The Criminal Investigations and Network Analysis Center (CINA) at George Mason University.

\ifCLASSOPTIONcaptionsoff
  \newpage
\fi

\bibliography{cite}

\begin{thebibliography}{10}

\bibitem{handbook}
D.~Maltoni, D.~Maio, A.~K. Jain, and S.~Prabhakar, {\em Handbook of fingerprint
  recognition}.
\newblock Springer Science \& Business Media, 2009.

\bibitem{fvc_ongoing}
B.~Dorizzi, R.~Cappelli, M.~Ferrara, D.~Maio, D.~Maltoni, N.~Houmani,
  S.~Garcia-Salicetti, and A.~Mayoue, ``Fingerprint and on-line signature
  verification competitions at icb 2009,'' in {\em International Conference on
  Biometrics}, pp.~725--732, Springer, 2009.

\bibitem{deepprint}
J.~J. Engelsma, K.~Cao, and A.~K. Jain, ``Learning a fixed-length fingerprint
  representation,'' {\em IEEE Transactions on Pattern Analysis and Machine
  Intelligence}, 2019.

\bibitem{engelsma2020hers}
J.~J. Engelsma, A.~K. Jain, and V.~N. Boddeti, ``Hers: Homomorphically
  encrypted representation search,'' {\em arXiv preprint arXiv:2003.12197},
  2020.

\bibitem{index1}
K.~Cao and A.~K. Jain, ``Fingerprint \uppercase{i}ndexing and
  \uppercase{m}atching: An \uppercase{i}ntegrated \uppercase{a}pproach,'' in
  {\em Biometrics (IJCB), 2017 IEEE International Joint Conference on},
  pp.~437--445, IEEE, 2017.

\bibitem{index2}
D.~Song and J.~Feng, ``Fingerprint \uppercase{i}ndexing based on
  \uppercase{p}yramid \uppercase{d}eep \uppercase{c}onvolutional
  \uppercase{f}eature,'' in {\em Biometrics (IJCB), 2017 IEEE International
  Joint Conference on}, pp.~200--207, IEEE, 2017.

\bibitem{index3}
D.~Song, Y.~Tang, and J.~Feng, ``Aggregating
  \uppercase{m}inutia-\uppercase{c}entred \uppercase{d}eep
  \uppercase{c}onvolutional \uppercase{f}eatures for \uppercase{f}ingerprint
  \uppercase{i}ndexing,'' {\em Pattern Recognition}, vol.~88, pp.~397--408,
  2019.

\bibitem{index4}
R.~Li, D.~Song, Y.~Liu, and J.~Feng, ``Learning \uppercase{G}lobal
  \uppercase{F}ingerprint \uppercase{F}eatures by \uppercase{T}raining a
  \uppercase{F}ully \uppercase{C}onvolutional \uppercase{N}etwork with
  \uppercase{L}ocal \uppercase{P}atches,'' {\em to appear in IEEE ICB}, 2019.

\bibitem{longitudinal}
S.~Yoon and A.~K. Jain, ``Longitudinal \uppercase{S}tudy of
  \uppercase{f}ingerprint \uppercase{r}ecognition,'' {\em Proceedings of the
  National Academy of Sciences}, vol.~112, no.~28, pp.~8555--8560, 2015.

\bibitem{nist302}
G.~P. Fiumara, P.~A. Flanagan, J.~D. Grantham, K.~Ko, K.~Marshall, M.~Schwarz,
  E.~Tabassi, B.~Woodgate, and C.~Boehnen, ``Nist special database 302: Nail to
  nail fingerprint challenge,'' 2019.

\bibitem{fvc2002}
``Fvc2002.'' \url{http://bias.csr.unibo.it/fvc2002/}.
\newblock 2002.

\bibitem{fvc2004}
``Fvc2004.'' \url{http://bias.csr.unibo.it/fvc2004/}.
\newblock 2004.

\bibitem{livdet2019}
G.~Orr{\`u}, R.~Casula, P.~Tuveri, C.~Bazzoni, G.~Dessalvi, M.~Micheletto,
  L.~Ghiani, and G.~L. Marcialis, ``Livdet in action-fingerprint liveness
  detection competition 2019,'' in {\em 2019 International Conference on
  Biometrics (ICB)}, pp.~1--6, IEEE, 2019.

\bibitem{sd4}
C.~I. Watson and C.~L. Wilson, ``Nist special database 4,'' {\em Fingerprint
  Database, National Institute of Standards and Technology}, vol.~17, no.~77,
  p.~5, 1992.

\bibitem{sd14}
C.~I. Watson, ``Nist special database 14,'' {\em Fingerprint Database, US
  National Institute of Standards and Technology}, 1993.

\bibitem{sd27}
M.~D. Garris and M.~D. Garris, {\em NIST special database 27: Fingerprint
  minutiae from latent and matching tenprint images}.
\newblock US Department of Commerce, National Institute of Standards and
  Technology, 2000.

\bibitem{cappelli2002synthetic}
R.~Cappelli, D.~Maio, and D.~Maltoni, ``Synthetic fingerprint-database
  generation,'' in {\em Object recognition supported by user interaction for
  service robots}, vol.~3, pp.~744--747, IEEE, 2002.

\bibitem{zhao2012fingerprint}
Q.~Zhao, A.~K. Jain, N.~G. Paulter, and M.~Taylor, ``Fingerprint image
  synthesis based on statistical feature models,'' in {\em 2012 IEEE Fifth
  International Conference on Biometrics: Theory, Applications and Systems
  (BTAS)}, pp.~23--30, IEEE, 2012.

\bibitem{johnson2013texture}
P.~Johnson, F.~Hua, and S.~Schuckers, ``Texture modeling for synthetic
  fingerprint generation,'' in {\em Proceedings of the IEEE Conference on
  Computer Vision and Pattern Recognition Workshops}, pp.~154--159, 2013.

\bibitem{bontrager2018deepmasterprints}
P.~Bontrager, A.~Roy, J.~Togelius, N.~Memon, and A.~Ross, ``Deepmasterprints:
  Generating masterprints for dictionary attacks via latent variable
  evolution,'' in {\em 2018 IEEE 9th International Conference on Biometrics
  Theory, Applications and Systems (BTAS)}, pp.~1--9, IEEE, 2018.

\bibitem{finger-gan}
S.~Minaee and A.~Abdolrashidi, ``Finger-gan: Generating realistic fingerprint
  images using connectivity imposed gan,'' {\em arXiv preprint
  arXiv:1812.10482}, 2018.

\bibitem{attia2019fingerprint}
M.~Attia, M.~H. Attia, J.~Iskander, K.~Saleh, D.~Nahavandi, A.~Abobakr,
  M.~Hossny, and S.~Nahavandi, ``Fingerprint synthesis via latent space
  representation,'' in {\em 2019 IEEE International Conference on Systems, Man
  and Cybernetics (SMC)}, pp.~1855--1861, IEEE, 2019.

\bibitem{synfi}
M.~S. Riazi, S.~M. Chavoshian, and F.~Koushanfar, ``Synfi: Automatic synthetic
  fingerprint generation,'' {\em arXiv preprint arXiv:2002.08900}, 2020.

\bibitem{lightweight}
M.~A.-N.~I. Fahim and H.~Y. Jung, ``A lightweight gan network for large scale
  fingerprint generation,'' {\em IEEE Access}, vol.~8, pp.~92918--92928, 2020.

\bibitem{kai_synthetic}
K.~{Cao} and A.~{Jain}, ``Fingerprint synthesis: Evaluating fingerprint search
  at scale,'' in {\em 2018 International Conference on Biometrics (ICB)}, 2018.

\bibitem{mistry2019fingerprint}
V.~Mistry, J.~J. Engelsma, and A.~K. Jain, ``Fingerprint synthesis: Search with
  100 million prints,'' in {\em 2020 International Joint Conference on
  Biometrics (IJCB)}, 2020.

\bibitem{level-3}
A.~B.~V. Wyzykowski, M.~P. Segundo, and R.~d.~P. Lemes, ``Level three synthetic
  fingerprint generation,'' {\em arXiv preprint arXiv:2002.03809}, 2020.

\bibitem{bahmani2021high}
K.~Bahmani, R.~Plesh, P.~Johnson, S.~Schuckers, and T.~Swyka, ``High fidelity
  fingerprint generation: Quality, uniqueness, and privacy,'' {\em arXiv
  preprint arXiv:2105.10403}, 2021.

\bibitem{msceleb}
Y.~Guo, L.~Zhang, Y.~Hu, X.~He, and J.~Gao, ``Ms-celeb-1m: A dataset and
  benchmark for large-scale face recognition,'' in {\em European conference on
  computer vision}, pp.~87--102, Springer, 2016.

\bibitem{vgg_face}
Q.~Cao, L.~Shen, W.~Xie, O.~M. Parkhi, and A.~Zisserman, ``Vggface2: A dataset
  for recognising faces across pose and age,'' in {\em International Conference
  on Automatic Face and Gesture Recognition}, 2018.

\bibitem{charles}
D.~Wang, C.~Otto, and A.~K. Jain, ``Face search at scale,'' {\em IEEE
  transactions on pattern analysis and machine intelligence}, vol.~39, no.~6,
  pp.~1122--1136, 2016.

\bibitem{tinsley2021face}
P.~Tinsley, A.~Czajka, and P.~Flynn, ``This face does not exist... but it might
  be yours! identity leakage in generative models,'' in {\em Proceedings of the
  IEEE/CVF Winter Conference on Applications of Computer Vision},
  pp.~1320--1328, 2021.

\bibitem{feng2021gans}
Q.~Feng, C.~Guo, F.~Benitez-Quiroz, and A.~M. Martinez, ``When do gans
  replicate? on the choice of dataset size,'' in {\em Proceedings of the
  IEEE/CVF International Conference on Computer Vision}, pp.~6701--6710, 2021.

\bibitem{nagarajan2018theoretical}
V.~Nagarajan, C.~Raffel, and I.~J. Goodfellow, ``Theoretical insights into
  memorization in gans,'' in {\em Neural Information Processing Systems
  Workshop}, vol.~1, 2018.

\bibitem{cappelli_iet}
R.~Cappelli, M.~Ferrara, and D.~Maltoni, ``{Generating synthetic
  fingerprints},'' in {\em {Hand-Based Biometrics: Methods and technology, IET,
  2018}}.

\bibitem{nfiq}
E.~Tabassi, ``{NFIQ 2.0: NIST Fingerprint image quality},'' {\em NISTIR 8034},
  2016.

\bibitem{sherlock1993model}
B.~G. Sherlock and D.~M. Monro, ``A model for interpreting fingerprint
  topology,'' {\em Pattern recognition}, vol.~26, no.~7, pp.~1047--1055, 1993.

\bibitem{larkin2007coherent}
K.~G. Larkin and P.~A. Fletcher, ``A coherent framework for fingerprint
  analysis: are fingerprints holograms?,'' {\em Optics Express}, vol.~15,
  no.~14, pp.~8667--8677, 2007.

\bibitem{fogel1989gabor}
I.~Fogel and D.~Sagi, ``Gabor filters as texture discriminator,'' {\em
  Biological cybernetics}, vol.~61, no.~2, pp.~103--113, 1989.

\bibitem{chen2014svm}
S.~Chen, S.~Chang, Q.~Huang, J.~He, H.~Wang, and Q.~Huang, ``Svm-based
  synthetic fingerprint discrimination algorithm and quantitative optimization
  strategy,'' {\em PloS one}, vol.~9, no.~10, p.~e111099, 2014.

\bibitem{gottschlich2014separating}
C.~Gottschlich and S.~Huckemann, ``Separating the real from the synthetic:
  minutiae histograms as fingerprints of fingerprints,'' {\em IET Biometrics},
  vol.~3, no.~4, pp.~291--301, 2014.

\bibitem{brock2018large}
A.~Brock, J.~Donahue, and K.~Simonyan, ``Large scale gan training for high
  fidelity natural image synthesis,'' {\em arXiv preprint arXiv:1809.11096},
  2018.

\bibitem{ulyanov2016instance}
D.~Ulyanov, A.~Vedaldi, and V.~Lempitsky, ``Instance normalization: The missing
  ingredient for fast stylization,'' {\em arXiv preprint arXiv:1607.08022},
  2016.

\end{thebibliography}


\begin{thebibliography}{10}

\bibitem{cao2023comprehensive}
Y.~Cao, S.~Li, Y.~Liu, Z.~Yan, Y.~Dai, P.~S. Yu, and L.~Sun, ``A comprehensive survey of ai-generated content (aigc): A history of generative ai from gan to chatgpt,'' {\em arXiv preprint arXiv:2303.04226}, 2023.

\bibitem{openai2022chatgpt}
OpenAI, ``Chatgpt.'' \url{https://openai.com/blog/chatgpt}, October 2022.

\bibitem{google-imagen}
Google, ``Introducing imagen: A large-scale image library and open-source tooling to enable computer vision research,'' May 2021.

\bibitem{sora_openai}
OpenAI, ``Creating video from text,'' {\em OpenAI Blog}, February 2024.

\bibitem{goodfellow2014generative}
I.~Goodfellow, J.~Pouget-Abadie, M.~Mirza, B.~Xu, D.~Warde-Farley, S.~Ozair, A.~Courville, and Y.~Bengio, ``Generative adversarial nets,'' {\em Advances in neural information processing systems}, vol.~27, 2014.

\bibitem{dhariwal2021diffusion}
P.~Dhariwal and A.~Nichol, ``Diffusion models beat gans on image synthesis,'' {\em Advances in neural information processing systems}, vol.~34, pp.~8780--8794, 2021.

\bibitem{cappelli2004sfinge}
R.~Cappelli, D.~Maio, and D.~Maltoni, ``Sfinge: an approach to synthetic fingerprint generation,'' in {\em International Workshop on Biometric Technologies}, pp.~147--154, 2004.

\bibitem{mistry2019fingerprint}
V.~Mistry, J.~J. Engelsma, and A.~K. Jain, ``Fingerprint synthesis: Search with 100 million prints,'' in {\em International Joint Conference on Biometrics}, 2020.

\bibitem{finger-gan}
S.~Minaee and A.~Abdolrashidi, ``Finger-gan: Generating realistic fingerprint images using connectivity imposed gan,'' {\em arXiv preprint arXiv:1812.10482}, 2018.

\bibitem{synfi}
M.~S. Riazi, S.~M. Chavoshian, and F.~Koushanfar, ``Synfi: Automatic synthetic fingerprint generation,'' {\em arXiv preprint arXiv:2002.08900}, 2020.

\bibitem{lightweight}
M.~A.~I. Fahim and H.~Y. Jung, ``A lightweight gan network for large scale fingerprint generation,'' {\em IEEE Access}, vol.~8, pp.~92918--92928, 2020.

\bibitem{bahmani2021high}
K.~Bahmani, R.~Plesh, P.~Johnson, S.~Schuckers, and T.~Swyka, ``High fidelity fingerprint generation: Quality, uniqueness, and privacy,'' in {\em 2021 IEEE International Conference on Image Processing (ICIP)}, pp.~3018--3022, IEEE, 2021.

\bibitem{cao2018fingerprint}
K.~Cao and A.~Jain, ``Fingerprint synthesis: Evaluating fingerprint search at scale,'' in {\em 2018 International Conference on Biometrics (ICB)}, pp.~31--38, IEEE, 2018.

\bibitem{level-3}
A.~B.~V. Wyzykowski, M.~P. Segundo, and R.~de~Paula~Lemes, ``Level three synthetic fingerprint generation,'' in {\em 2020 25th International Conference on Pattern Recognition (ICPR)}, pp.~9250--9257, IEEE, 2021.

\bibitem{engelsma2022printsgan}
J.~J. Engelsma, S.~A. Grosz, and A.~K. Jain, ``Printsgan: synthetic fingerprint generator,'' {\em IEEE Transactions on Pattern Analysis and Machine Intelligence}, 2022.

\bibitem{grosz2022spoofgan}
S.~A. Grosz and A.~K. Jain, ``Spoofgan: Synthetic fingerprint spoof images,'' {\em IEEE Transactions on Information Forensics and Security}, 2022.

\bibitem{shoshan2024fpgan}
A.~Shoshan, N.~Bhonker, E.~Ben~Baruch, O.~Nizan, I.~Kviatkovsky, J.~Engelsma, M.~Aggarwal, and G.~Medioni, ``Fpgan-control: A controllable fingerprint generator for training with synthetic data,'' in {\em Proceedings of the IEEE/CVF Winter Conference on Applications of Computer Vision}, pp.~6067--6076, 2024.

\bibitem{li2023diffusion}
K.~Li and X.~Yang, ``Diffusion probabilistic model based end-to-end latent fingerprint synthesis,'' in {\em 2023 IEEE 4th International Conference on Pattern Recognition and Machine Learning (PRML)}, pp.~343--349, IEEE, 2023.

\bibitem{tang2024enhancing}
W.~Tang, D.~Figueroa, D.~Liu, K.~Johnsson, and A.~Sopasakis, ``Enhancing fingerprint image synthesis with gans, diffusion models, and style transfer techniques,'' {\em arXiv preprint arXiv:2403.13916}, 2024.

\bibitem{rombach2022high}
R.~Rombach, A.~Blattmann, D.~Lorenz, P.~Esser, and B.~Ommer, ``High-resolution image synthesis with latent diffusion models,'' in {\em Proceedings of the IEEE/CVF conference on computer vision and pattern recognition}, pp.~10684--10695, 2022.

\bibitem{von-platen-etal-2022-diffusers}
P.~von Platen, S.~Patil, A.~Lozhkov, P.~Cuenca, N.~Lambert, K.~Rasul, M.~Davaadorj, D.~Nair, S.~Paul, W.~Berman, Y.~Xu, S.~Liu, and T.~Wolf, ``Diffusers: State-of-the-art diffusion models.'' \url{https://github.com/huggingface/diffusers}, 2022.

\bibitem{tabassi2021nfiq}
E.~Tabassi, M.~Olsen, O.~Bausinger, C.~Busch, A.~Figlarz, G.~Fiumara, O.~Henniger, J.~Merkle, T.~Ruhland, C.~Schiel, {\em et~al.}, ``Nfiq 2 nist fingerprint image quality,'' 2021.

\bibitem{hu2022lora}
E.~J. Hu, Y.~Shen, P.~Wallis, Z.~Allen-Zhu, Y.~Li, S.~Wang, L.~Wang, and W.~Chen, ``Lo{RA}: Low-rank adaptation of large language models,'' in {\em International Conference on Learning Representations}, 2022.

\bibitem{loshchilov2016sgdr}
I.~Loshchilov and F.~Hutter, ``Sgdr: Stochastic gradient descent with warm restarts,'' {\em arXiv preprint arXiv:1608.03983}, 2016.

\bibitem{kingma2014adam}
D.~P. Kingma and J.~Ba, ``Adam: A method for stochastic optimization,'' {\em arXiv preprint arXiv:1412.6980}, 2014.

\bibitem{sd14}
C.~I. Watson, ``Nist special database 14: Mated fingerprint cards pairs 2 version 2,'' {\em tech. rep., Citeseer}, 2001.

\bibitem{fvc2002}
D.~Maio, D.~Maltoni, R.~Cappelli, J.~L. Wayman, and A.~K. Jain, ``Fvc2002: Second fingerprint verification competition,'' in {\em 2002 International Conference on Pattern Recognition}, vol.~3, pp.~811--814, IEEE, 2002.

\bibitem{fvc2004}
D.~Maio, D.~Maltoni, R.~Cappelli, J.~L. Wayman, and A.~K. Jain, ``Fvc2004: Third fingerprint verification competition,'' in {\em Biometric Authentication: First International Conference, ICBA 2004, Hong Kong, China, July 15-17, 2004. Proceedings}, pp.~1--7, Springer, 2004.

\bibitem{kirchgasser2021plus}
S.~Kirchgasser, C.~Kauba, and A.~Uhl, ``The plus multi-sensor and longitudinal fingerprint dataset: An initial quality and performance evaluation,'' {\em IEEE Transactions on Biometrics, Behavior, and Identity Science}, vol.~4, no.~1, pp.~43--56, 2021.

\bibitem{j2019infant}
J.~J~Engelsma, D.~Deb, A.~Jain, A.~Bhatnagar, and P.~Sewak~Sudhish, ``Infant-prints: Fingerprints for reducing infant mortality,'' in {\em Proceedings of the IEEE/CVF Conference on Computer Vision and Pattern Recognition Workshops}, pp.~67--74, 2019.

\bibitem{nist302}
G.~P. Fiumara, P.~A. Flanagan, J.~D. Grantham, K.~Ko, K.~Marshall, M.~Schwarz, E.~Tabassi, B.~Woodgate, and C.~Boehnen, ``Nist special database 302: Nail to nail fingerprint challenge,'' Tech. Rep. NIST.TN.2007, National Institute of Standards and Technology, Gaithersburg, MD, 2019.

\bibitem{yoon2015longitudinal}
S.~Yoon and A.~K. Jain, ``Longitudinal study of fingerprint recognition,'' {\em Proceedings of the National Academy of Sciences}, vol.~112, no.~28, pp.~8555--8560, 2015.

\bibitem{sankaran2012hierarchical}
A.~Sankaran, M.~Vatsa, and R.~Singh, ``Hierarchical fusion for matching simultaneous latent fingerprint,'' in {\em 2012 IEEE Fifth International Conference on Biometrics: Theory, Applications and Systems (BTAS)}, pp.~377--382, IEEE, 2012.

\bibitem{sankaran2015multisensor}
A.~Sankaran, M.~Vatsa, and R.~Singh, ``Multisensor optical and latent fingerprint database,'' {\em IEEE access}, vol.~3, pp.~653--665, 2015.

\bibitem{malhotra2023multi}
A.~Malhotra, M.~Vatsa, R.~Singh, K.~B. Morris, and A.~Noore, ``Multi-surface multi-technique (must) latent fingerprint database,'' {\em IEEE Transactions on Information Forensics and Security}, 2023.

\bibitem{birajadar2019towards}
P.~Birajadar, M.~Haria, P.~Kulkarni, S.~Gupta, P.~Joshi, B.~Singh, and V.~Gadre, ``Towards smartphone-based touchless fingerprint recognition,'' {\em S{\=a}dhan{\=a}}, vol.~44, no.~7, pp.~1--15, 2019.

\bibitem{malhotra2020matching}
A.~Malhotra, A.~Sankaran, M.~Vatsa, and R.~Singh, ``On matching finger-selfies using deep scattering networks,'' {\em IEEE Transactions on Biometrics, Behavior, and Identity Science}, vol.~2, no.~4, pp.~350--362, 2020.

\bibitem{zhou2014benchmark}
W.~Zhou, J.~Hu, I.~Petersen, S.~Wang, and M.~Bennamoun, ``A benchmark 3d fingerprint database,'' in {\em 2014 11th International Conference on Fuzzy Systems and Knowledge Discovery (FSKD)}, pp.~935--940, IEEE, 2014.

\bibitem{grosz2021c2cl}
S.~A. Grosz, J.~J. Engelsma, E.~Liu, and A.~K. Jain, ``C2cl: Contact to contactless fingerprint matching,'' {\em IEEE Transactions on Information Forensics and Security}, vol.~17, pp.~196--210, 2021.

\bibitem{simonyan2014very}
K.~Simonyan and A.~Zisserman, ``Very deep convolutional networks for large-scale image recognition,'' {\em arXiv preprint arXiv:1409.1556}, 2014.

\bibitem{johnson2016perceptual}
J.~Johnson, A.~Alahi, and L.~Fei-Fei, ``Perceptual losses for real-time style transfer and super-resolution,'' in {\em European Conference on Computer Vision, 2016}, pp.~694--711, Springer, 2016.

\bibitem{gal2022image}
R.~Gal, Y.~Alaluf, Y.~Atzmon, O.~Patashnik, A.~H. Bermano, G.~Chechik, and D.~Cohen-Or, ``An image is worth one word: Personalizing text-to-image generation using textual inversion,'' {\em arXiv preprint arXiv:2208.01618}, 2022.

\bibitem{ruiz2023dreambooth}
N.~Ruiz, Y.~Li, V.~Jampani, Y.~Pritch, M.~Rubinstein, and K.~Aberman, ``Dreambooth: Fine tuning text-to-image diffusion models for subject-driven generation,'' in {\em Proceedings of the IEEE/CVF Conference on Computer Vision and Pattern Recognition}, pp.~22500--22510, 2023.

\bibitem{wang2024instantid}
Q.~Wang, X.~Bai, H.~Wang, Z.~Qin, and A.~Chen, ``Instantid: Zero-shot identity-preserving generation in seconds,'' {\em arXiv preprint arXiv:2401.07519}, 2024.

\bibitem{li2023photomaker}
Z.~Li, M.~Cao, X.~Wang, Z.~Qi, M.-M. Cheng, and Y.~Shan, ``Photomaker: Customizing realistic human photos via stacked id embedding,'' {\em arXiv preprint arXiv:2312.04461}, 2023.

\bibitem{zhang2023adding}
L.~Zhang, A.~Rao, and M.~Agrawala, ``Adding conditional control to text-to-image diffusion models,'' in {\em Proceedings of the IEEE/CVF International Conference on Computer Vision}, pp.~3836--3847, 2023.

\bibitem{grosz2023latent}
S.~A. Grosz and A.~K. Jain, ``Latent fingerprint recognition: Fusion of local and global embeddings,'' {\em IEEE Transactions on Information Forensics and Security}, 2023.

\bibitem{liu2024latent}
X.~Liu, K.~Raja, R.~Wang, H.~Qiu, H.~Wu, D.~Sun, Q.~Zheng, N.~Liu, X.~Wang, G.~Huang, {\em et~al.}, ``A latent fingerprint in the wild database,'' {\em IEEE Transactions on Information Forensics and Security}, 2024.

\bibitem{grosz2022afr}
S.~A. Grosz and A.~K. Jain, ``Afr-net: Attention-driven fingerprint recognition network,'' {\em IEEE Transactions on Biometrics, Behavior, and Identity Science}, 2023.

\bibitem{van2008visualizing}
L.~Van~der Maaten and G.~Hinton, ``Visualizing data using t-sne.,'' {\em Journal of machine learning research}, vol.~9, no.~11, 2008.

\bibitem{he2016deep}
K.~He, X.~Zhang, S.~Ren, and J.~Sun, ``Deep residual learning for image recognition,'' in {\em Proceedings of the IEEE conference on computer vision and pattern recognition}, pp.~770--778, 2016.

\bibitem{dosovitskiy2020image}
A.~Dosovitskiy, L.~Beyer, A.~Kolesnikov, D.~Weissenborn, X.~Zhai, T.~Unterthiner, M.~Dehghani, M.~Minderer, G.~Heigold, S.~Gelly, {\em et~al.}, ``An image is worth 16x16 words: Transformers for image recognition at scale,'' {\em arXiv preprint arXiv:2010.11929}, 2020.

\bibitem{lin2018cnn}
C.~Lin and A.~Kumar, ``A cnn-based framework for comparison of contactless to contact-based fingerprints,'' {\em IEEE Transactions on Information Forensics and Security}, vol.~14, no.~3, pp.~662--676, 2018.

\bibitem{sd4}
C.~I. Watson and C.~L. Wilson, ``Nist special database 4,'' {\em Fingerprint Database, National Institute of Standards and Technology}, vol.~17, no.~77, p.~5, 1992.

\bibitem{sd27}
M.~D. Garris, {\em NIST special database 27: Fingerprint minutiae from latent and matching tenprint images}.
\newblock US Department of Commerce, National Institute of Standards and Technology, 2000.

\bibitem{boddeti2023biometric}
V.~N. Boddeti, G.~Sreekumar, and A.~Ross, ``On the biometric capacity of generative face models,'' in {\em 2023 IEEE International Joint Conference on Biometrics (IJCB)}, pp.~1--10, IEEE, 2023.

\end{thebibliography}
\bibliographystyle{ieeetr}

\begin{IEEEbiography}
[{\includegraphics[width=1in,height=1.25in,clip,keepaspectratio]{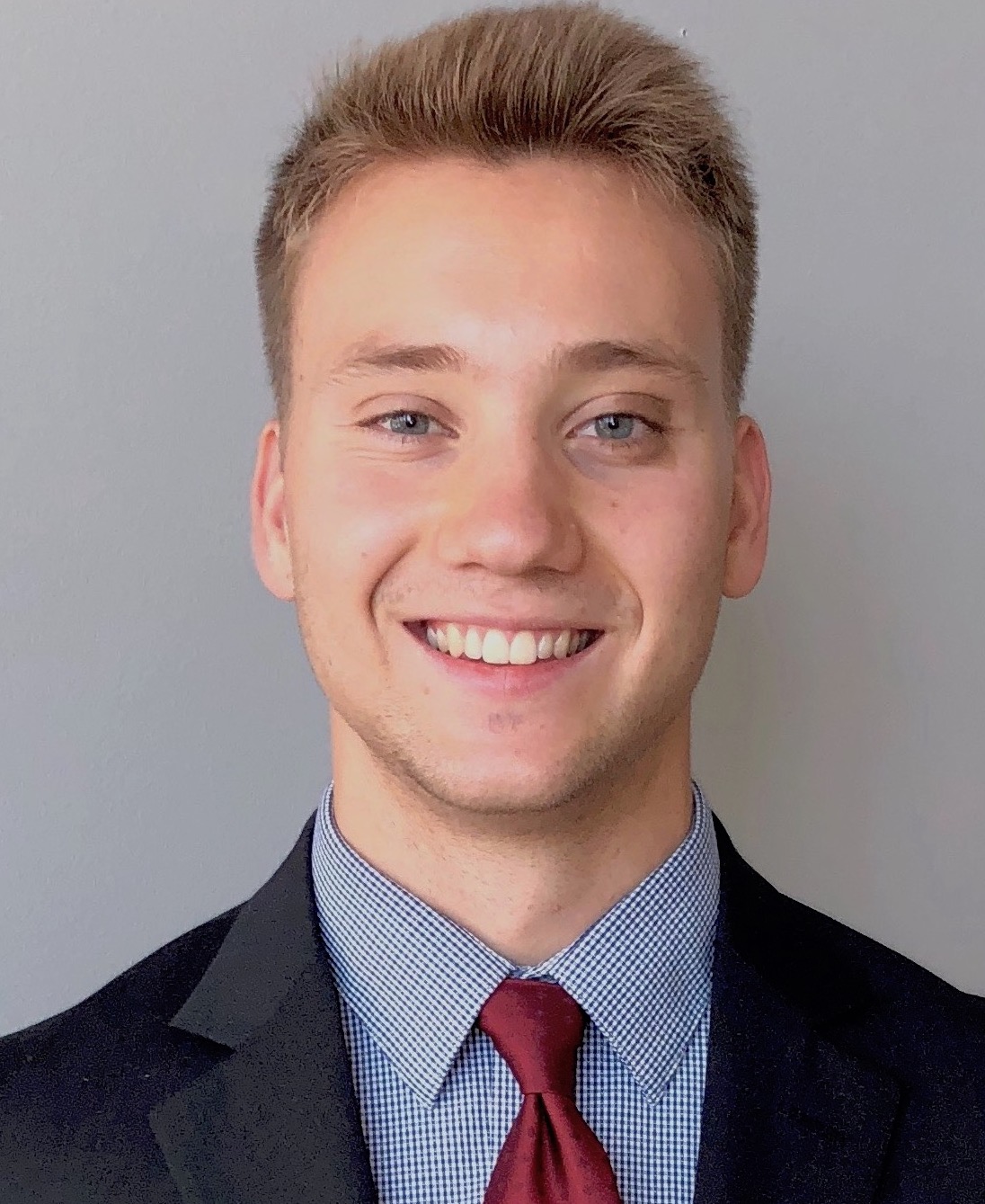}}]{Steven A. Grosz}
received his B.S. degree with highest honors in Electrical Engineering from Michigan State University, East Lansing, Michigan, in 2019. He is currently a doctoral student in the Department of Computer Science and Engineering at Michigan State University. His primary research interests are in the areas of machine learning and computer vision with applications in biometrics.
\vspace{-1.5em}
\end{IEEEbiography}

\begin{IEEEbiography}[{\includegraphics[width=1in,height=1.25in,clip,keepaspectratio]{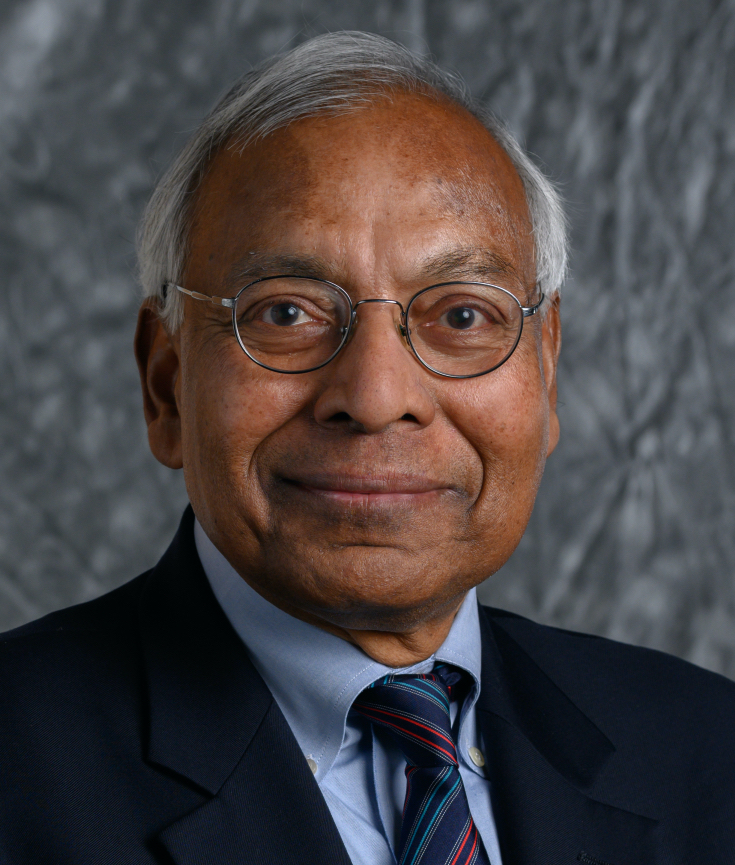}}]{Anil K. Jain}
Anil K. Jain is a University distinguished professor in the Department of Computer Science and Engineering at Michigan State University. His research interests include pattern recognition, computer vision, and biometric authentication. He served as the editor-in-chief of the IEEE Transactions on Pattern Analysis and Machine Intelligence and was a member of the United States Defense Science Board. He has received Fulbright, Guggenheim, Alexander von Humboldt, and IAPR King Sun Fu awards. He is a member of the National Academy of Engineering, the Indian National Academy of Engineering, the World Academy of Sciences, and the Chinese Academy of Sciences.
\vspace{-1.5em}
\end{IEEEbiography}

\end{document}